\documentclass[review]{elsarticle}
\graphicspath{ {./figures/} }
\usepackage{hyperref}
\usepackage{float}
\usepackage{verbatim} 
\usepackage{apalike}

\usepackage{subcaption}
\usepackage{graphicx}
\usepackage{colortbl}
\usepackage{multirow}
\usepackage{tabularx}
	\newcolumntype{L}{>{\raggedright\arraybackslash}X}
	\newcolumntype{Z}{>{\centering\let\newline\\\arraybackslash\hspace{0pt}}X}
	\newcolumntype{C}{>{\centering\arraybackslash}X}
	\newcolumntype{K}[1]{>{\centering\arraybackslash}p{#1}}

\definecolor{Gray}{gray}{0.85}
\definecolor{LightCyan}{rgb}{0.58,1,1}
\restylefloat{figure}
\restylefloat{table}

\let\today\relax
\makeatletter
\def\ps@pprintTitle{%
    \let\@oddhead\@empty
    \let\@evenhead\@empty
    \def\@oddfoot{\footnotesize\itshape
         {Preprint submitted to Expert Systems with Applications} \hfill}%
    \let\@evenfoot\@oddfoot
    }
\makeatother

\journal{Expert Systems with Applications}

\biboptions{authoryear}

\begin{document}
\begin{frontmatter}

\title{Automatic Vision-Based Parking Slot Detection and Occupancy Classification}

\author[label1]{Ratko Grbi\'{c} \corref{cor1}}
\ead{ratko.grbic@ferit.hr}

\author[label1]{Brando Koch}
\ead{bkoch4142@gmail.com}

\cortext[cor1]{Corresponding author.}
\address[label1]{Faculty of Electrical Engineering, Computer Science and Information Technology Osijek, Kneza Trpimira 2B, Osijek, HR-31000, Croatia}

\begin{abstract}
Parking guidance information (PGI) systems are used to provide information to drivers about the nearest parking lots and the number of vacant parking slots. Recently, vision-based solutions started to appear as a cost-effective alternative to standard PGI systems based on hardware sensors mounted on each parking slot. Vision-based systems provide information about parking occupancy based on images taken by a camera that is recording a parking lot. However, such systems are challenging to develop due to various possible viewpoints, weather conditions, and object occlusions. Most notably, they require manual labeling of parking slot locations in the input image which is sensitive to camera angle change, replacement, or maintenance. In this paper, the algorithm that performs Automatic Parking Slot Detection and Occupancy Classification (APSD-OC) solely on input images is proposed. Automatic parking slot detection is based on vehicle detections in a series of parking lot images upon which clustering is applied in bird's eye view to detect parking slots. Once the parking slots positions are determined in the input image, each detected parking slot is classified as occupied or vacant using a specifically trained ResNet34 deep classifier. The proposed approach is extensively evaluated on well-known publicly available datasets (PKLot and CNRPark+EXT), showing high efficiency in parking slot detection and robustness to the presence of illegal parking or passing vehicles. Trained classifier achieves high accuracy in parking slot occupancy classification.
\end{abstract}


\end{frontmatter}

\section{Introduction}\label{sec:Introduction}

Urban population and the number of motor vehicles are constantly increasing and thus saturating not only the road network but also space reserved for the vehicles parking. The shortage of parking space is especially pronounced in crowded urban areas where the available parking space is in high demand throughout the whole day. With limited parking space, the drivers are forced to cruise for vacant parking slots and thus are creating mobile queues which affect normal traffic flow. Apart from traffic congestion, cruising for parking space creates additional pollution and CO\textsubscript{2} emissions, negatively impacts driving time, and creates additional costs for the driver \cite{Zhu2020}.

To make it easier to find a vacant parking space and to enable more efficient utilization of available parking spaces, different parking guidance information (PGI) systems are installed by parking providers \cite{Lin2017}. Such systems require accurate and real-time information about the occupancy of each individual parking slot in a certain parking area in order to provide relevant information to the nearby drivers via mobile apps or information panels \cite{Guo2014}. A straightforward approach for getting information about the occupancy of each parking slot is based on the installation of some kind of sensor on each parking slot. Typically, ultrasonic sensor \cite{Chen2011a}, magnetic sensor \cite{ZushengZhang2015,Sifuentes2011} or even the combination of two sensors \cite{SENSIT,Alam2018} is mounted on each parking slot to detect the presence of a vehicle. While such approaches can provide highly accurate information to a PGI system, they require additional costs in terms of sensors cost, installation, and maintenance. Recently, parking slot occupancy detection based on computer vision emerged as a promising source of information for PGI systems \cite{Cai2019,Acharya2018,Ng2020,Amato2017}. These solutions extract information about parking occupancy from images obtained by a camera that is recording parking area from a certain viewpoint such that parking slots are at least partially visible. This approach can be attractive since it does not require any additional sensors besides the camera and can provide additional information regarding parking usage such as vehicle recognition. Since many parking areas are already covered with surveillance cameras, such approach to parking occupancy detection can be quite cost-effective with respect to sensor-based systems and can be implemented quickly and more easily. Obviously, bad weather conditions like snow or rain, significant illuminance changes during daytime and nighttime operation, different camera viewpoints, vehicle occlusion, and presence of passing and illegally parked vehicles are the main challenges when developing and implementing vision-based PGI systems.

A significant problem in most of the vision-based solutions for parking slot occupancy classification is the requirement for manual labeling of the parking slots in the input image obtained by a camera that is recording a certain parking area, i.e. a parking lot. The resulting annotations are then used to extract every individual parking lot from the camera image. Each extracted patch is then classified as occupied or not by using some kind of classifier. This classifier can follow traditional computer vision approach (typically feature extraction + SVM classifier \cite{DeAlmeida2015}) or can be based on deep learning (DL) approach \cite{Acharya2018,Amato2017}. However, this manual labeling can be cumbersome and time-consuming if the system is going to process images from several cameras. Apart from that, in case of camera angle change or zooming, maintenance, or camera replacement the labeling procedure must be repeated. In this spirit, automatic parking space detection is recently pointed out in \cite{ALMEIDA2022} as a significant problem of modern vision-based parking lot management.

A bunch of training data is a must if DL classifier is to be used for parking slot occupancy classification. This lack of a consistent and representative dataset was recognized by the research community. In \cite{DeAlmeida2015}, the PKLot dataset was proposed which is a relatively large dataset that contains labeled images from two parking scenes from different camera views in different weather conditions. Similarly, in \cite{Amato2017} CNRPark+EXT dataset is presented which contains annotated images of parking scenes in different weather conditions as well during the night. Both datasets are publicly available and are often used for the training and benchmarking of DL classifiers.

There are several attempts reported in literature which are using object detection as a source of information regarding parking slot occupancy. In \cite{Ke2020} authors are using single-stage object detector on edge devices and send the detections for further analysis by the PGI system server. In \cite{Padmasiri2020} automated vehicle parking occupancy detection is performed in real-time with two-stage detectors Faster R-CNN and RetinaNet. However, many false positive detections arise in such approach due to passing vehicles. Additionally, most of such approaches are focused on the implementation part without getting into detailed evaluation on publicly available datasets like \cite{DeAlmeida2015,Amato2017} and some require additional a priori knowledge to efficiently judge parking occupancy status based on input image \cite{MartinNieto2019}.

In this paper, vision-based algorithm for Automatic Parking Slot Detection and Occupancy Classification (APSD-OC) is proposed. The proposed APSD-OC removes the need for manual labeling of parking slots and further improves the accuracy of the occupancy classification. We approach the problem of parking slots position determination as a vehicle detection problem in the series of images captured by a camera through a certain period of time and by taking into account that drivers usually park their vehicles inside marked parking slots. The vehicle detection is performed on a set of input images followed by the appropriate algorithm for determining positions of the marked parking slots which includes perspective transformation and detections clustering. As such, APSD-OC can distinguish regular parking slots from parking violations, i.e. vehicles parked outside any marked parking slots, and consequently exclude the latter from further analysis. Once the parking slots are detected, in each subsequently captured image these parking slots are cropped and each is classified as occupied or vacant with the proposed classifier. The complete procedure can be easily applied to any parking surveillance problem since it does not require any additional parameters which are difficult to obtain like the camera angle, camera parameters, or homography matrix. The only required parameter to be provided by end-user is the number of visible parking slots which can be easily obtained and thus the whole algorithm can be applied to a different parking lot relatively easily. The proposed approach is extensively analyzed on the two well-known publicly available datasets and is compared with the state-of-the-art solutions showing the high efficiency both in parking slot detection and parking slot occupancy classification.

The paper is structured as follows. In Section~\ref{sec:related_work} related work is overviewed, advantages and disadvantages of recent approaches are pointed out. The proposed algorithm for automatic parking slot detection and occupancy classification is presented in Section ~\ref{sec:methodology}. The description of used datasets and how experiments are performed are given in Section~\ref{sec:results_and_discussion} with the obtained results and accompanying discussion. In the end, conclusions are given with guidelines for future work.

\section{Related Work}
\label{sec:related_work}

A rough categorization can be made on vehicle-driven and space-driven methods as proposed in \cite{Huang2010}. The vehicle-driven methods are focused on vehicle detection upon which vacant parking slots are determined. In the latter case, the focus is on direct detection of the available parking slots in an overall scene. Similarly, in \cite{MartinNieto2019} a categorization is given regarding how methods perform parking slot classification: image segmentation based systems \cite{Huang2010,Al-Kharusi2014}, machine learning over parking slots patches \cite{7280319,Amato2017} and vehicle detection techniques based on object detectors \mbox{\cite{MartinNieto2019,Padmasiri2020,Xie2015}}.

Also, two main directions of research can be differentiated in the literature regarding used algorithms. The first one is based on traditional computer vision techniques \cite{Al-Kharusi2014} which are often coupled with (shallow) machine learning methods (e.g. support vector machines - SVM) to answer the question of whether a certain parking slot is occupied or not \cite{DeAlmeida2015}. The second direction of research is more recent and uses deep learning as its main engine to determine parking slot occupancy status \cite{Nurullayev2019,Acharya2018,Amato2017,Padmasiri2020,Ke2020,Cai2019,9352169,MartinNieto2019}. In both cases, a certain mechanism is implemented or a priori information is given to the system which determines actual parking slots in the scene. Some of the aforementioned papers are overviewed in more detail in the rest of this section.

An intelligent parking management system, based solely on traditional image processing techniques, is proposed in \cite{Al-Kharusi2014}. This includes colorspace transformation, morphological operation (dilate and erode), thresholding, edge detection, and Hough transform. The focus of the paper is on the overall system and the actual efficiency regarding parking slot classification is not reported. The authors also point out that the method does not have the same efficiency in different weather conditions.

Although the problem of parking occupancy can be solved with traditional computer vision techniques, most of the recent papers introduce machine learning to build a more efficient parking slot occupancy classifier. Paper \cite{DeAlmeida2015} is a significant work in the field of parking management since it proposed the PKLot dataset and thus enabled a systematic benchmark of different parking occupancy classification methods. Apart from that, the authors proposed a parking slot occupancy SVM classifier based upon textual descriptors such as Local Binary Patterns (LBP) and Local Phase Quantization (LPQ). When the same view of parking was used for training and testing the classifier, the recognition rate went over 99\%. However, when the classifier was trained on one set of parking lots and tested on another one, the recognition rate was 89\%, indicating that further research in generalization capabilities of used classifiers should be performed.

Many researchers recognized that hand-crafted visual features (SIFT, SURF, ORB, etc.) have a limited ability to adapt to variations of object appearance that are highly non-linear, time-varying, and complex. Interestingly, pretrained CNN showed as an excellent "off-the-shelf" feature extractor in many different visual recognition tasks \cite{Razavian2014}. Therefore, in \cite{Acharya2018} authors are using VGG CNN \cite{Simonyan2014} pretrained on ImageNet to efficiently extract features and to train a binary SVM classifier for the purpose of parking occupancy classification. The authors trained the classifier on the PKLot dataset and tested it on a specific dataset called Barry street. This dataset contains 810 images of the parking lot with 30 parking slots resulting in 24300 annotations, i.e. patches of occuppied or vacant parking slots. The proposed approach achieved 96.6\% accuracy on the Barry street dataset. The authors also point out that in practice it should be able to detect the predefined areas of the parking slots automatically rather than manually identifying the boundaries.
 
In \cite{Ahrnbom2016} authors developed a parking slot occupancy classifier that combines Integral Channel Features (ICF) with Logistic Regression (LR) or SVM. The proposed method was designed to achieve good accuracy and robustness but keeping in mind overall method complexity so that it can run on embedded devices. At first, ten feature channels are extracted for each input image, such as color channels in LUV color space, gradient magnitude, and quantized gradient channels. Then, feature vectors are calculated from a certain feature channel in an efficient manner using the integral image approach. In the end, logistic regression and SVM classifiers are trained. Both types of classifiers are trained and tested on the PKLot dataset.

Authors in \cite{Amato2017} specifically designed a deep neural network called mAlexNet for the purpose of parking occupancy classification. The efficiency of the method is extensively tested on PKLot and CNRPark-EXT datasets and outperforms AlexNet and LPQ from \cite{DeAlmeida2015} in terms of classification accuracy and area under the curve (AUC). Interestingly, this deep network is three orders of magnitude smaller than the original AlexNet and can be implemented on an embedded platform like Raspberry Pi 2 model B.

In \cite{Nurullayev2019}, CarNet is proposed which is DNN that uses dilated convolutional neural network to indicate parking space occupancy status. Input to the CarNet is a 54x32 RGB image of a parking slot. The presented experiments show that CarNet outperforms AlexNet \cite{Krizhevsky2012} and other well-known DL architectures on PKLot dataset and mAlexNet \cite{Amato2017} on CNRPark-EXT dataset. While CarNet achieves quite high precision and robustness, it requires that parking slot images are manually cropped from the input image of a whole parking lot. The comparison of CarNet with our approach can be found in the results section.

Authors in \cite{Ke2020} point out that a key component of modern smart cities is traffic surveillance which needs significant computing power and storage. In case of a parking space occupancy, computing workload can be moved toward the edge, i.e. local devices equipped with cameras. The proposed approach balances computational load and data transmission volume. Therefore, a single-stage object detector called Single Shot Detector (SSD) is used on edge devices to detect vehicles. The detections are sent to the server which runs object tracking (to reduce false positive detections) and occupancy judgment algorithms. The obtained results on a parking garage use-case show that such approach yields efficient and reliable detection performance in various environmental conditions.

In \cite{Padmasiri2020}, a scalable software architecture solution is proposed which enables reliable implementation of an end-to-end automated vehicle parking occupancy detection system. The vehicle detection is performed in real-time with two-stage detectors Faster R-CNN and RetinaNet. The proposed approach is tested on the PKLot dataset and the efficiency of the object detector is expressed in average precision (AP). The status of each parking lot is obtained by an object detection algorithm but the actual parking slots position detection is missing so that moving vehicles are often detected as occupied parking spaces. The authors put significant effort into system usability so the developed web-based and mobile-based applications enable end-users easy finding of free parking slots.

Video-based parking occupancy detection is proposed in \cite{Chen2020}. Hereby, YOLOv3 \cite{redmon2018yolov3} based on MobileNet version 2 is used for vehicle detection. To train YOLOv3, authors have firstly labeled the CNRPark dataset. Additionally, the authors implemented a voting mechanism to prevent false positive classifications of parking spaces caused by large vehicles passing by. The solution was evaluated using the CNRPark-EXT dataset, a simulated model of a street with parking lots, and real images taken with a camera. The information regarding parking lot occupancy is further processed by a streetlight control system. The authors report high accuracy on the CNRPark+EXT dataset. Unfortunately, test data and annotations are not publicly available.

Obviously, manual labeling of the parking spots as proposed in \cite{Amato2017,Nurullayev2019} can be cumbersome and time-consuming, especially in the case of parking with large number of parking slots. Additionally, in the case of camera movement, the labeling process must be repeated. The need for automatical identification of the parking slots boundaries in practical applications is clearly pointed out in \cite{Acharya2018,ALMEIDA2022}. There are several attempts in the literature to perform automatic parking slot detections. In \cite{MartinNieto2019} authors proposed automatica detection of each parking slot when car park area is rectangular and forming a parking grid. However, additional information must be provided, e.g. user must specify the corners of the parking area and the number of slots, as well as camera homography matrix has to be estimated. Recently, two step automatic parking slot detection is proposed in \cite{Patel2020}. At first, vehicle detection is performed using Faster R-CNN or Yolov4 and then vehicle tracking is performed to distinguish between stationary vehicles and moving vehicles. While the proposed apporach obtains pretty high recall values for CNRPark-EXT datasets (even 100\% during busy days when all slots are occupied), it appears that false positives can occur due to illegally parked vehicles.

\section{Proposed Approach to Automatic Parking Slot Detection and Occupancy Classification}
\label{sec:methodology}

\begin{figure*}[tbh]
	\centering
  	\includegraphics[width=0.9\linewidth]{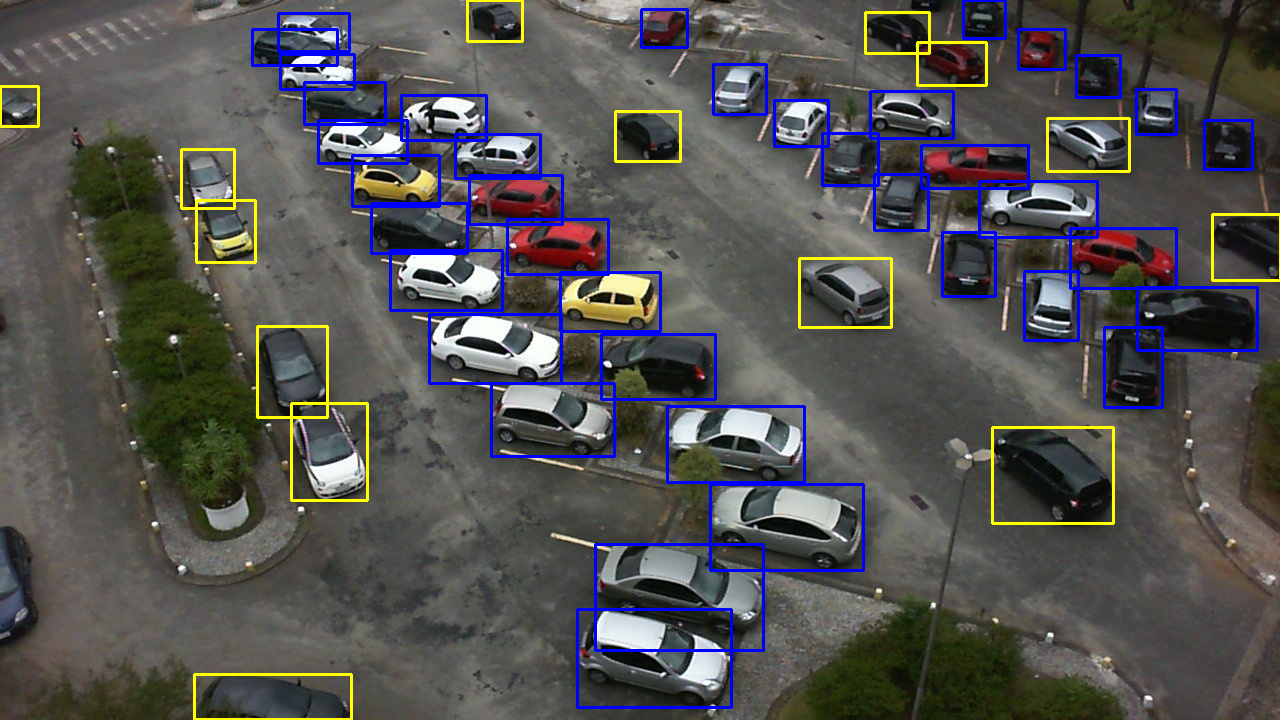}  
	\caption{An example of a parking lot image from PKLot dataset. Properly parked vehicles are marked with blue bounding box.}
	\label{fig:parking_example} 
\end{figure*}

A parking slot is defined as an area that is designed for a single vehicle parking and is usually marked with painted white or blue lines on a road surface. Automatic parking slot detection is defined as the localization of a parking slot within the input image obtained by a camera that is recording the parking area. One way to describe the parking slot location in the image is by using a rectangular bounding box (BB) which is defined by its center, width, and height in pixels. Our approach for automatic parking slot detection relies on the processing of the camera images which are acquired with a certain sampling frequency. This is typically 5 minutes in practice. The core assumption of our proposed approach is that parking slots are parts of an image where repeated vehicle detections occur since the drivers are forced to park the vehicle inside the marked area. However, one should be aware that drivers sometimes make parking violations, i.e. they park vehicles in a prohibited space such as entrance or sidewalk, and which can lead to congestion or even accidents. Such locations should be ignored in parking slot detection since they are not valid parking slots. An example of a parking lot image from the PKLot dataset is shown in Figure~\ref{fig:parking_example}. Properly parked vehicles are those parked inside slot markings and are marked with blue BBs. The rest of the vehicles are making parking violations and are marked with yellow BBs. Once the detection of parking slots is finished, newly acquired images are processed to obtain the status of each parking slot (is it occupied or vacant) and the information can be sent to a PGI system.

\begin{figure*}[tbh]
	\centering
  	\includegraphics[width=0.95\linewidth]{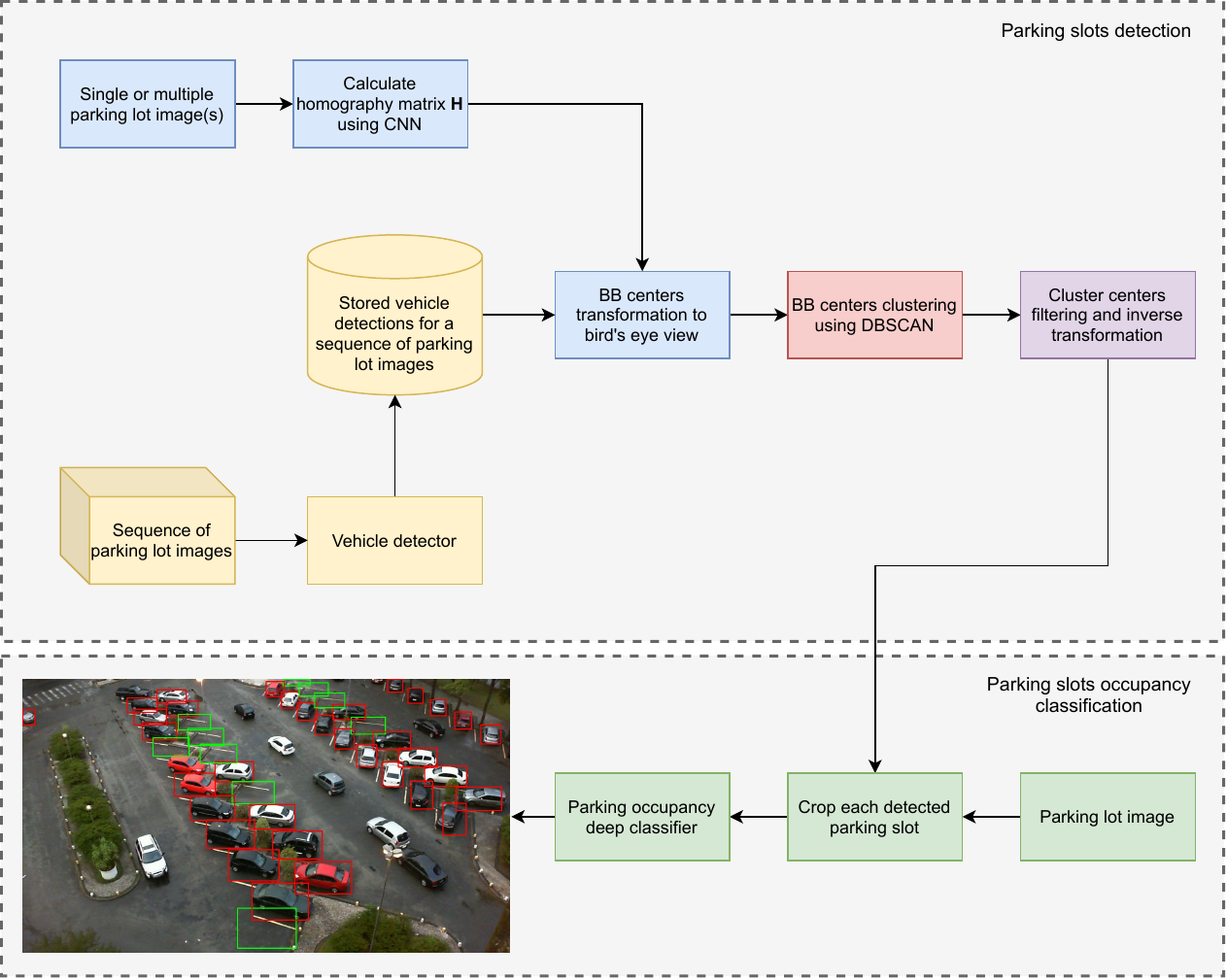}  
	\caption{The block diagram of the proposed APSD-OC algorithm.}
	\label{fig:APSD-OC} 
\end{figure*}

The proposed Automatic Parking Slot Detection and Occupancy Classification (APSD-OC) algorithm contains two main parts as can be seen in Figure~\ref{fig:APSD-OC} with grey overlays. The upper part performs parking slot detection outputting the corresponding rectangular area for each detected parking slot. Once the detection part is finished, the bottom part determines the status of each detected parking slot using a deep classifier.

Automatic parking slot detection is based on the processing of available parking lot images which were captured using a camera with a fixed position and angle. These images should be collected over a period of time in which all parking slots are used for vehicle parking so they can be properly detected. To each obtained image an object detector is applied which outputs a rectangular BB for each detected vehicle inside the image. Once all images are processed, BB centers clustering is performed to reveal locations where vehicles are detected consistently. However, since parking cameras can be mounted in different physical locations and angles, images can be obtained from different perspectives which can affect the efficiency of the clustering process due to different BB center densities in areas closer to the camera versus areas more far back. Therefore, BB centers are firstly transformed using a homography matrix which relates the original camera view and the bird's eye view of the parking lot. Once the cluster centers are obtained in bird's eye view, these are further examined by analyzing the distribution of BB centers that belong to each cluster. By doing so it can be distinguished to a certain degree if a cluster center corresponds to a regular parking slot or to some kind of unmarked parking lot area where sporadical or illegal parking occurs. Parking slot occupancy classification can be relatively easily performed once the precise locations of parking slots are available. All detected parking slots are cropped from the input image and are processed using a deep classifier which outputs the probability that each parking slot is occupied.

\subsection{Vehicle Detection Using YOLOv5}
\label{subsec:vehicle_detection}

\begin{figure*}[tbh]
\centering
\begin{subfigure}[t]{.45\textwidth}
  \centering
  \includegraphics[width=\linewidth]{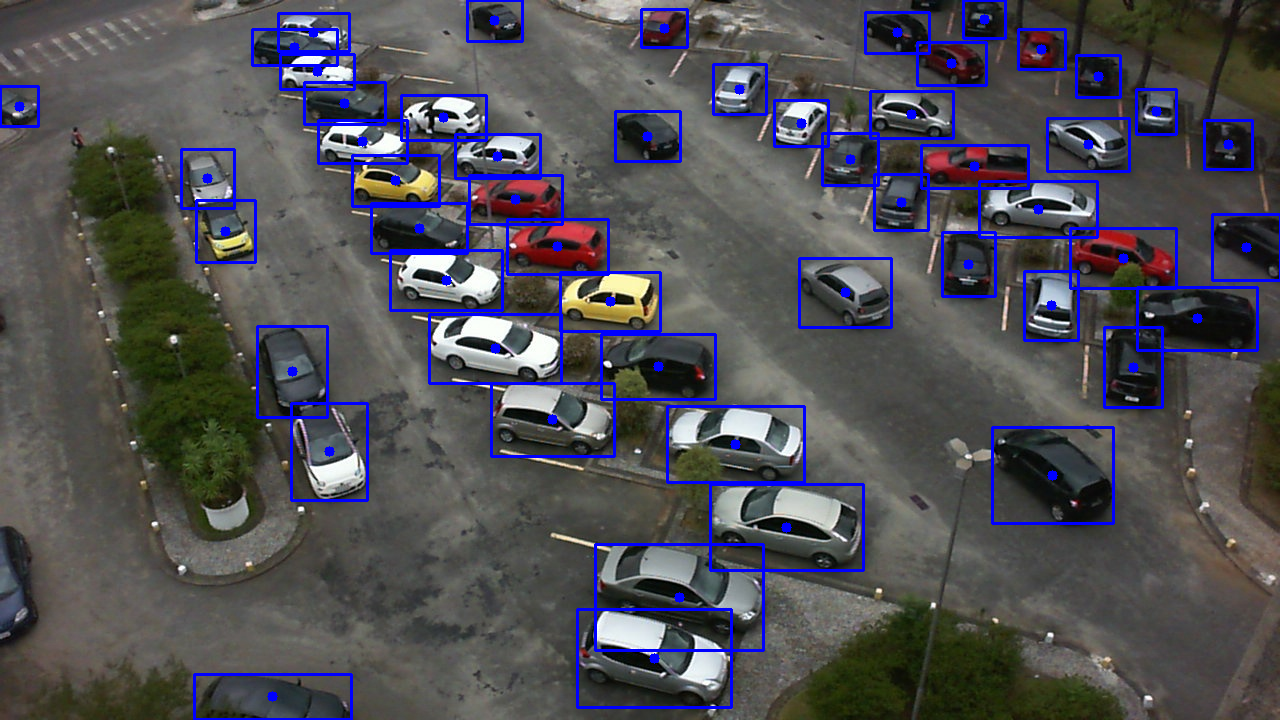}  
  \caption{Vehicle detection in a single input image.}
  \label{fig:vehicle_detection}
\end{subfigure}
\begin{subfigure}[t]{.45\textwidth}
  \centering
  \includegraphics[width=\linewidth]{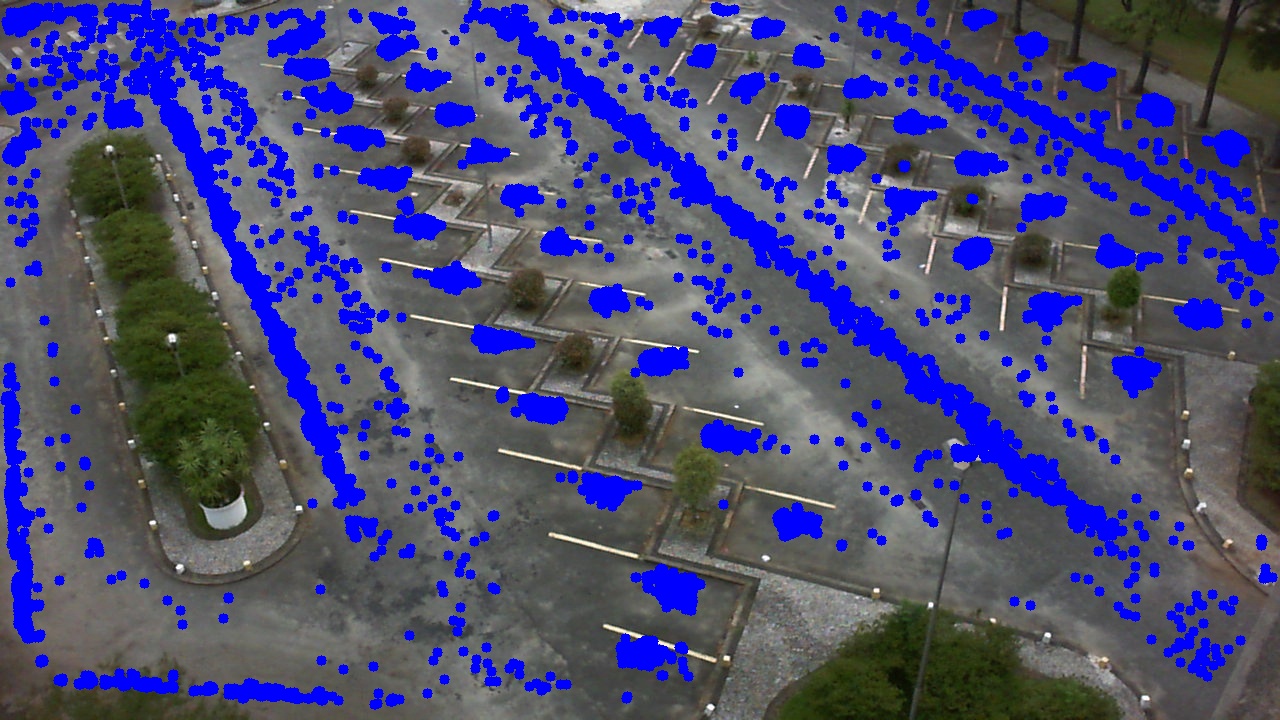}  
  \caption{Vehicle detection in multiple input images.}
  \label{fig:vehicle_detection_multiple}
\end{subfigure}
\hfill
\caption{Vehicle detection using Yolov5 in input images from PKLot dataset.}
\label{fig:detection_vehicles_examples} 
\end{figure*}

The first step in the proposed APSD-OC algorithm is to detect vehicles in a sequence of $N$ input images taken by a camera recording the same parking lot for a certain period of time. A vehicle detection implies locating a vehicle in the image, typically with the rectangular BB which tightly surrounds the vehicle. Therefore, each detected vehicle in image is represented with a vector $\mathbf{v}_i=[\mathbf{c}_i, w_i, h_i]$ where $\mathbf{c}_i=[x_i,y_i]$ is BB center, $w_i$ is BB width, and $h_i$ is BB height. An example of vehicle detection in an input image from PKLot dataset is shown in Figure~\ref{fig:vehicle_detection} where BBs are marked with blue color and blue dots represent corresponding BB centers.

For the task of vehicle detection, we use YOLOv5, an object detector from the YOLO family of end-to-end deep learning models designed for real-time object detection. This model was selected as it has proven to have solid performance on our dataset images. YOLO object detectors \cite{redmon2018yolov3}, compared to R-CNNs \cite{Ren2015}, replace the need for a region proposal step by splitting the input into a grid of cells, each containing a predefined number and shape of BBs over which an objectiveness score and a class prediction are made. Objectiveness score is the probability of a particular cell containing the object, while the class prediction is the probability distribution over the classes. This architecture choice enabled the efficiency of YOLO family. For the purpose of detecting vehicles in images a version yolov5x of the YOLOv5 model pretrained on the COCO dataset is used \cite{yolov5}. COCO dataset contains 200,000 labeled images with 1.5 million object instances and 80 object categories. During inference, we only considered classes "car" and "truck" and used a confidence threshold of $0.5$ during the processing of all images. Before entering the network images were resized to $1280\times1280$ px.

In this algorithm step, vehicle detection is performed for $N$ available images for a given camera as shown in Figure~\ref{fig:vehicle_detection_multiple} (just centers of BBs are shown). The resulting BBs are stored for further processing with a clustering algorithm in order to detect actual parking slots and to discard locations corresponding to illegal parking, bus stops, etc. 

\subsection{Perspective Transformation}
\label{subsec:perspective_transformation}

Generally speaking, a camera that is recording a parking lot should be mounted in such a way that all parking slots can be seen as much as possible, i.e. vehicle presence and its position can be clearly distinguished in the obtained image. Therefore, installation of the camera depends on several factors for a particular parking lot which include camera field-of-view, number of parking lots, presence of the trees or other structures, and so on. For example, the PKLot dataset \cite{DeAlmeida2015} is created using a low cost full high definition camera (Microsoft LifeCam, HD-5000) positioned at the top of a nearby building to minimize the occlusion between adjacent parked vehicles.

Figure~\ref{fig:vehicle_detection} shows a single input image from PKLot dataset with highlighted BBs and their centers using detector from Subsection~\ref{subsec:vehicle_detection}. If $N$ such images are processed using this detector, as shown in Figure~\ref{fig:vehicle_detection_multiple}, then it can be noticed that clusters of BB centers emerge on positions where vehicles are often parked. Due to the camera perspective, adjacent clusters that correspond to parking slots closer to the camera appear more distant than the adjacent clusters corresponding to parking slots that are further away from the camera. Therefore, to make the clustering process more efficient, the APSD-OC algorithm uses perspective transformation to obtain a bird's eye view (or top view) of a parking lot, minimizing the discrepancy between the distance of the adjacent clusters for any part of the parking lot. 

The aforementioned perspective transformation can be achieved with the usage of 3x3 homography matrix $\mathbf{H}$ which maps each BB center image coordinates $\mathbf{c}_i=[x_i,y_i]^\top$ to the bird's eye view $\mathbf{c}^{'}_{i}=[x^{'}_i,y^{'}_i]^\top$ according to $[x^{'}_i,y^{'}_i,1]^\top= \mathbf{H} [x_i,y_i,1]^\top$. In practice, a homography matrix is not known and has to be estimated. For example, in \cite{MartinNieto2019} the homography matrix for each camera is obtained using four points from each camera viewpoint and each point correspondence in an image extracted from a top view. The top view is in this case obtained from Google Earth. Obviously, such an approach requires certain manual work and requires additional information (top view) which sometimes might not be available.

To make the proposed APSD-OC algorithm more general we automate homography matrix $\mathbf{H}$ estimation by using the approach of obtaining bird's eye view from an image proposed in \cite{Abbas2019}. Hereby, homography is parametrized with only four parameters corresponding to the vertical vanishing point and ground plane vanishing line (horizon) in the image. These are regressed directly using a CNN. The CNN is trained on a large synthetic dataset which contains ground truth for the horizon line and the vertical vanishing point and which is built using CARLA simulator. The dataset is created by randomly changing camera height, field of view, roll and tilt angle thus obtaining different camera positions and orientations that can be found in practice.

Figure~\ref{fig:all_transformed_centers} shows the perspective transformation of the image shown in Figure~\ref{fig:vehicle_detection_multiple}. It can be noticed that the distance between BB centers that correspond to adjacent parking slots is approximately the same in the whole image when the perspective transformation is applied, i.e. in bird's eye view.

\subsection{Bounding Box Centers Clustering Using DBSCAN}
\label{subsec:clustering_DBSCAN}

\begin{figure*}[tbh]
\centering
\begin{subfigure}[t]{.45\textwidth}
  \centering
  \includegraphics[width=\linewidth, trim=0 327px 0 0, clip]{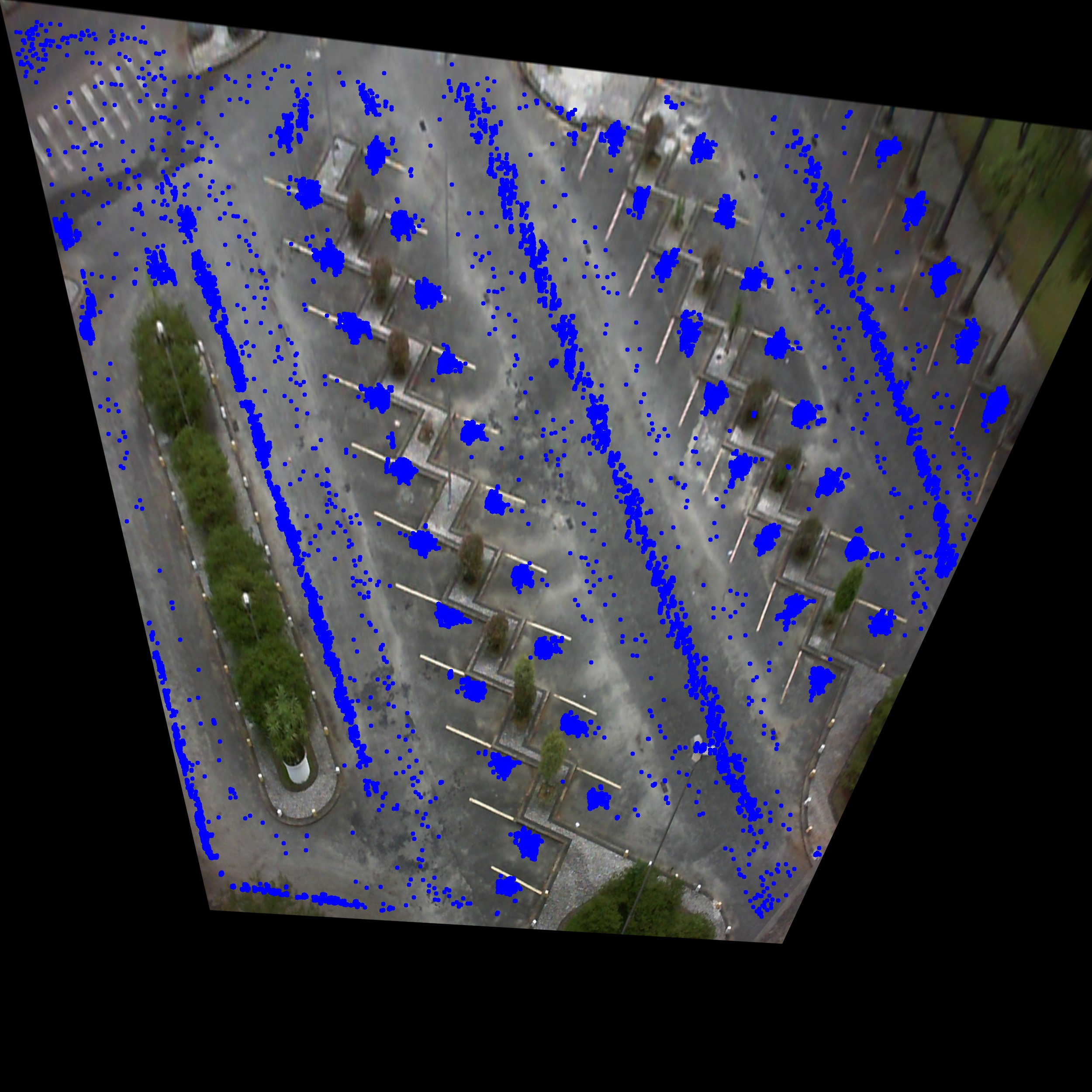}  
  \caption{Transformed BBs centers of vehicle detections for $N$ input images.}
  \label{fig:all_transformed_centers}
\end{subfigure}
\begin{subfigure}[t]{.45\textwidth}
  \centering
  \includegraphics[width=\linewidth, trim=0 327px 0 0, clip]{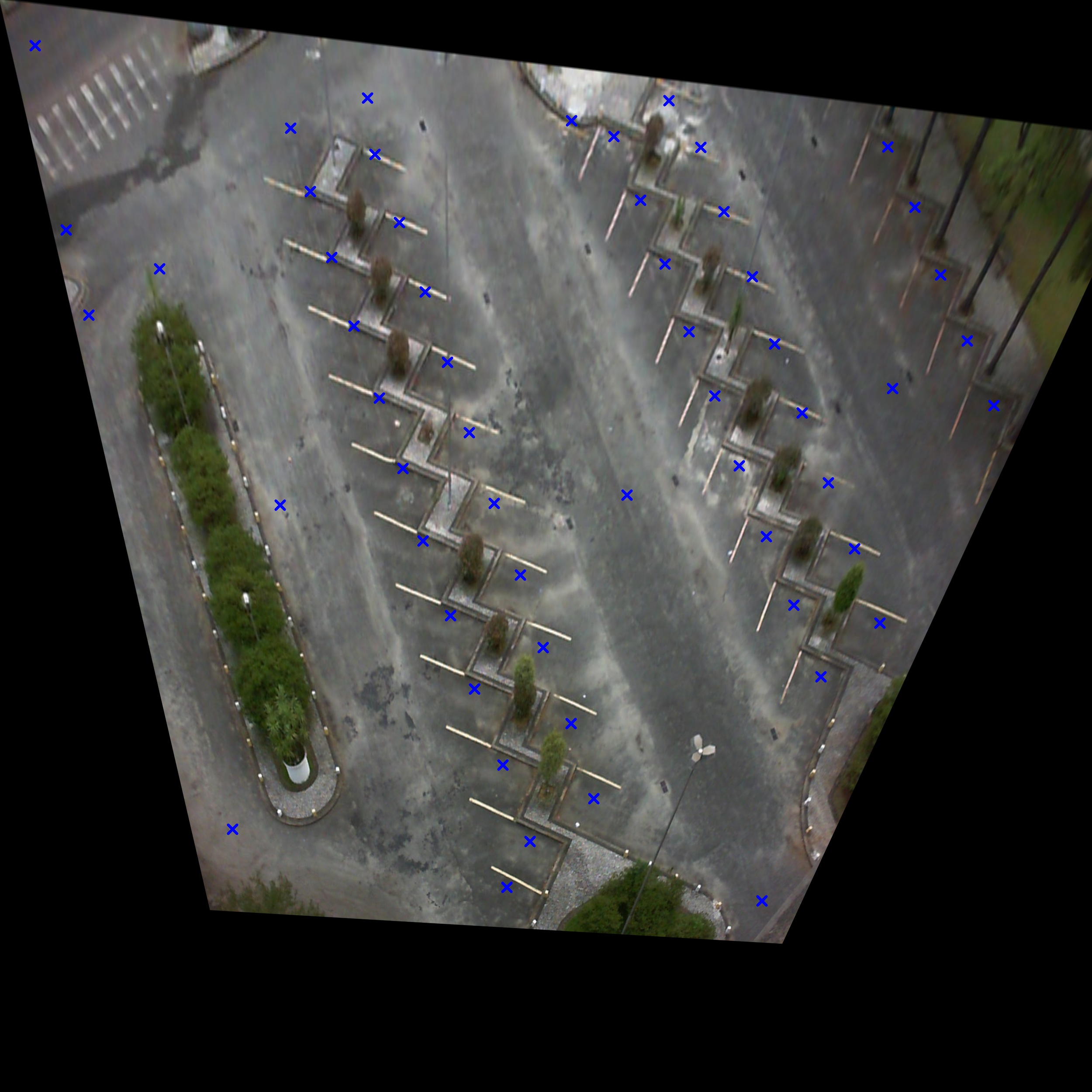}  
  \caption{Centers of clusters obtained by DBSCAN algorithm applied to transformed BB centers.}
  \label{fig:dbscan_clusters}
\end{subfigure}
\par\bigskip
\begin{subfigure}[t]{.45\textwidth}
  \centering
  \includegraphics[width=\linewidth, trim=0 327px 0 0, clip]{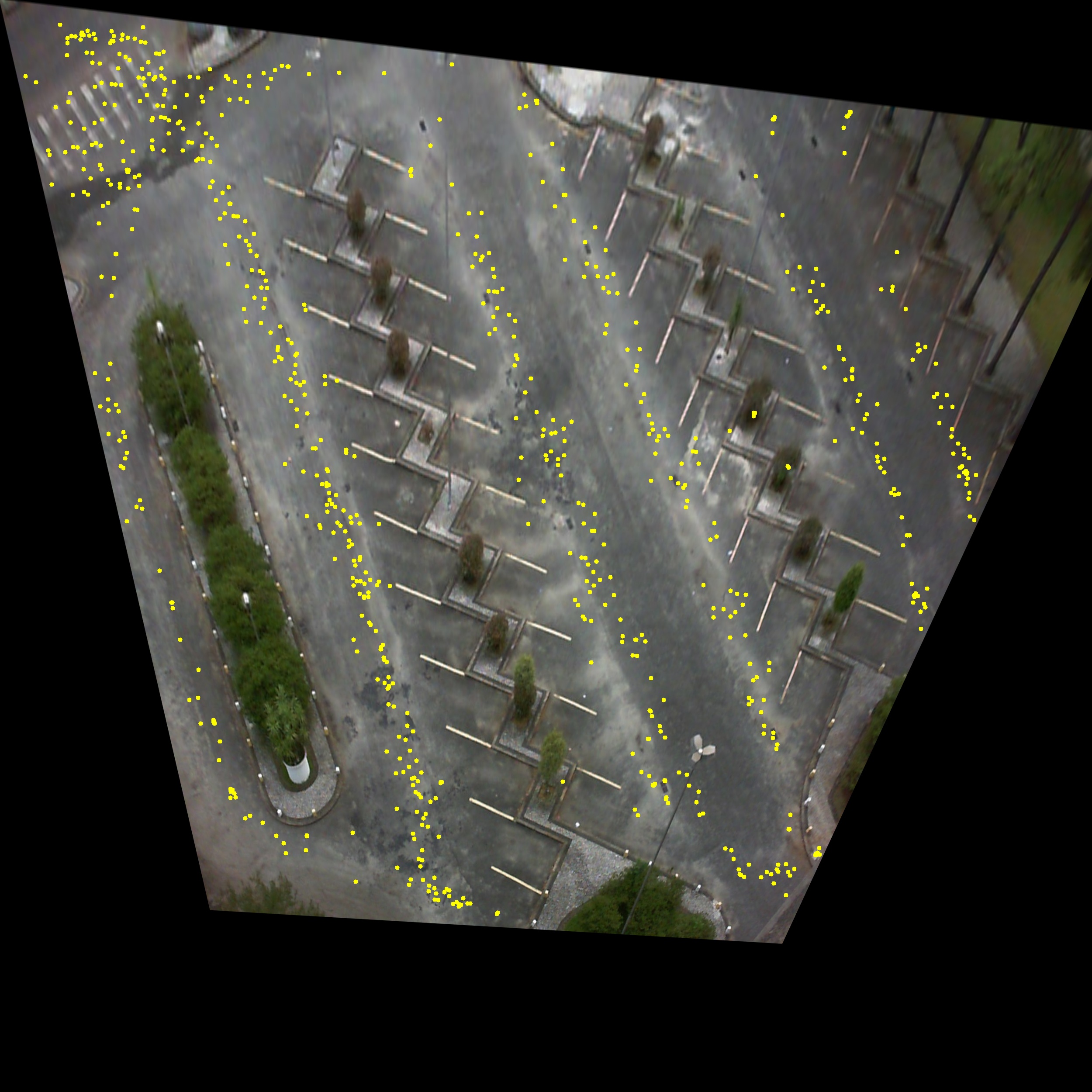}  
  \caption{Noise points of DBSCAN algorithm.\newline}
  \label{fig:dbscan_noise}
\end{subfigure}
\begin{subfigure}[t]{.45\textwidth}
  \centering
  \includegraphics[width=\linewidth, trim=0 327px 0 0, clip]{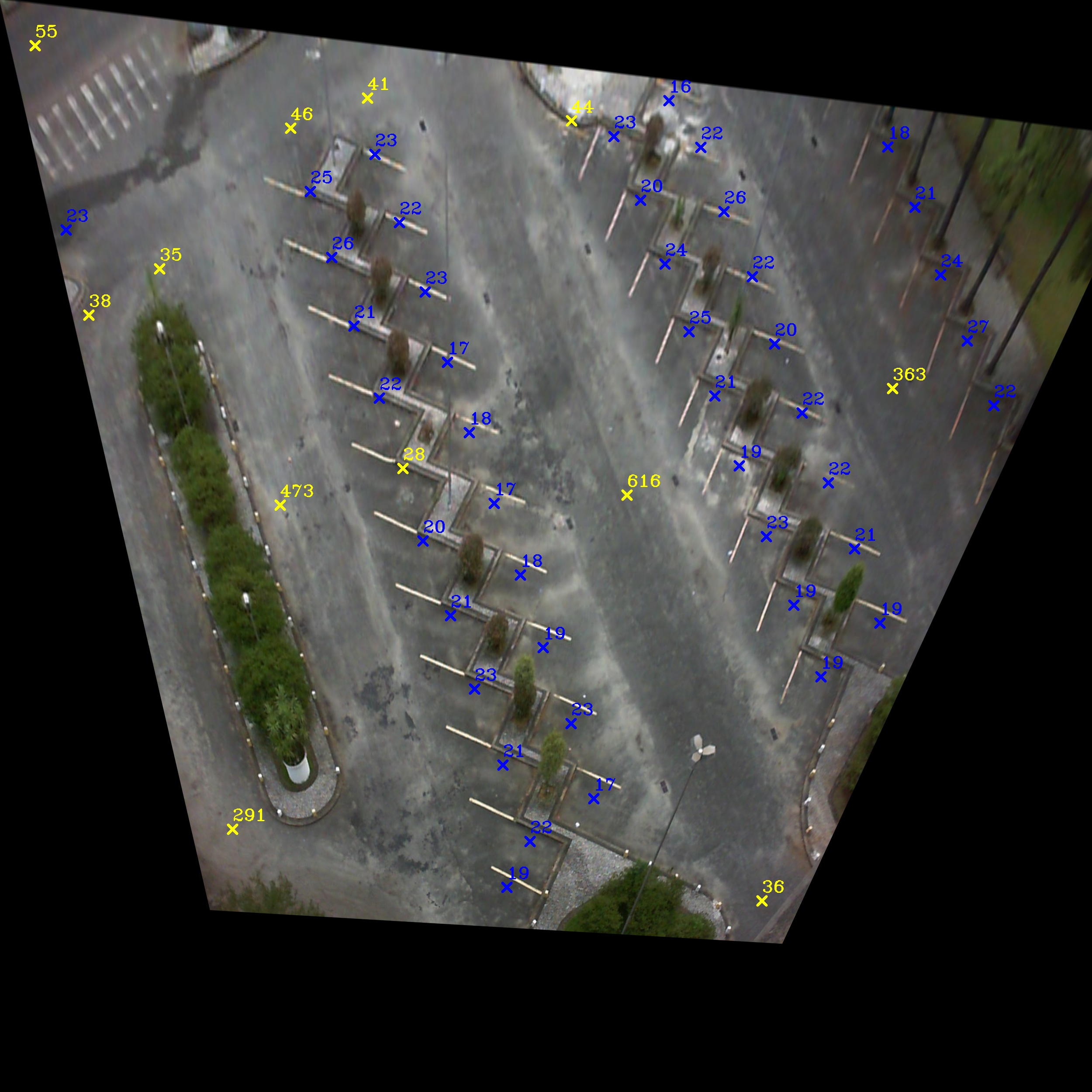}  
  \caption{Standard deviation $s_i$ of BBs centers around the corresponding cluster center $\mathbf{m}_i$.}
  \label{fig:cluster_BB_std}
\end{subfigure}
\caption{Example of clustering of BB centers using DBSCAN in bird's eye view - an example from PKLot dataset (UFPR05 camera).}
\label{fig:dbscan_clustering_BB_centers} 
\end{figure*}

Figure~\ref{fig:vehicle_detection_multiple} shows the image of the parking lot with overlayed vehicle detections in $N$ images and the corresponding bird's eye view is shown in Figure~\ref{fig:all_transformed_centers} (for the sake of simplicity just BB centers are shown). It can be noticed that BB centers tightly group around the center of each parking slot, but also there are a number of detections outside any parking markings which are due to illegal vehicle parking, passing vehicles, or even false vehicle detections. The former ones look like high-density regions and the latter ones appear as low-density regions (or outlying observations). In this case, it is beneficial to use some kind of density based spatial clustering in the bird's eye view.

Here we opt for a well-known DBSCAN algorithm \cite{dbscan}. DBSCAN relies on density based notion of clusters and it can detect clusters of arbitrary shape. Given the set of points, DBSCAN groups together points that are close to each other in certain space (usually Euclidean distance is applied) and if their count exceeds a predefined threshold. As such, DBSCAN requires only two parameters: $eps$ defines how close two points should be to each other to be considered as a part of the same cluster and $minPoints$ which is the minimum number of points for a region to be considered as a cluster. Points that are not reachable from any other point belonging a cluster are considered as outliers or noise points.

Figure~\ref{fig:dbscan_clusters} shows the application of DBSCAN algorithm on transformed BB centers from Figure~\ref{fig:all_transformed_centers}. Each cluster of BB centers is represented with mean value $\mathbf{m}_i$ of belonging points in bird's eye view. It can be noticed that most cluster centers $\mathbf{m}_i$ correspond to parking slots locations. However, some centers can be found in the middle of the road representing clusters that are formed by the detections of illegaly parked vehicles. Noise detections, or more precisely centers of such vehicle detections, are shown in Figure~\ref{fig:dbscan_noise}. Most of these noise points are due to passing cars or very short illegal parking (e.g. short stop behind a properly parked vehicle). For the next step of the algorithm only cluster centers $\mathbf{m}_i$ are considered while the noise points are discarded.

\subsection{Cluster Centers Filtering and Inverse Transformation}
\label{subsec:clusters_center_filtering}

The last step in automatic parking slot detection is filtering of the obtained cluster centers $\mathbf{m}_i$ and their transformation to the original view. Most of $\mathbf{m}_i$ values correspond to parking slots, see for example Figure~\ref{fig:dbscan_clusters}. However, some of these values can correspond to physical locations where parking violations often occur. To efficiently filter out such values, we take into account that the deviation of BBs centers around the corresponding mean values is usually much greater in case of parking violations than the deviation of BBs centers around the corresponding mean value for regular parking since in the latter case drivers are forced to park the vehicle inside slot markings. More precisely, for each $\mathbf{m}_i$ we define $s_i$ as the sum of standard deviations of $x$ and $y$ coordinates of corresponding BB centers. These $s_i$ values are then sorted in ascending order and every value outside the range $[Q1-1.5*IQR,Q3+1.5*IQR]$ is discarded. Hereby, $Q1$ is the first quartile of the BB centers, $Q3$ is the third quartile of the BB centers, and Inter-Quartile Range is $IQR=Q3-Q1$. After that, the first $n_{bottom}$ values are selected as final parking slots where $n_{bottom}$ corresponds to the number of visible parking slots and must be provided by the user. 

This step is illustrated in Figure~\ref{fig:cluster_BB_std} where each cluster center $\mathbf{m}_i$ has a corresponding standard deviation $s_i$. It can be noticed that cluster centers $\mathbf{m}_i$ corresponding to parking slots have a low value of $s_i$ while these values are significantly larger in locations where parking violations occur. For this particular parking lot the parameter $n_{bottom}$ is equal to $44$. The cluster centers $\mathbf{m}_i$, which are filtered out in this algorithm step, are shown in yellow color.

\begin{figure*}[!htb]
\centering
\begin{subfigure}[t]{.45\textwidth}
  \centering
  \includegraphics[width=\linewidth]{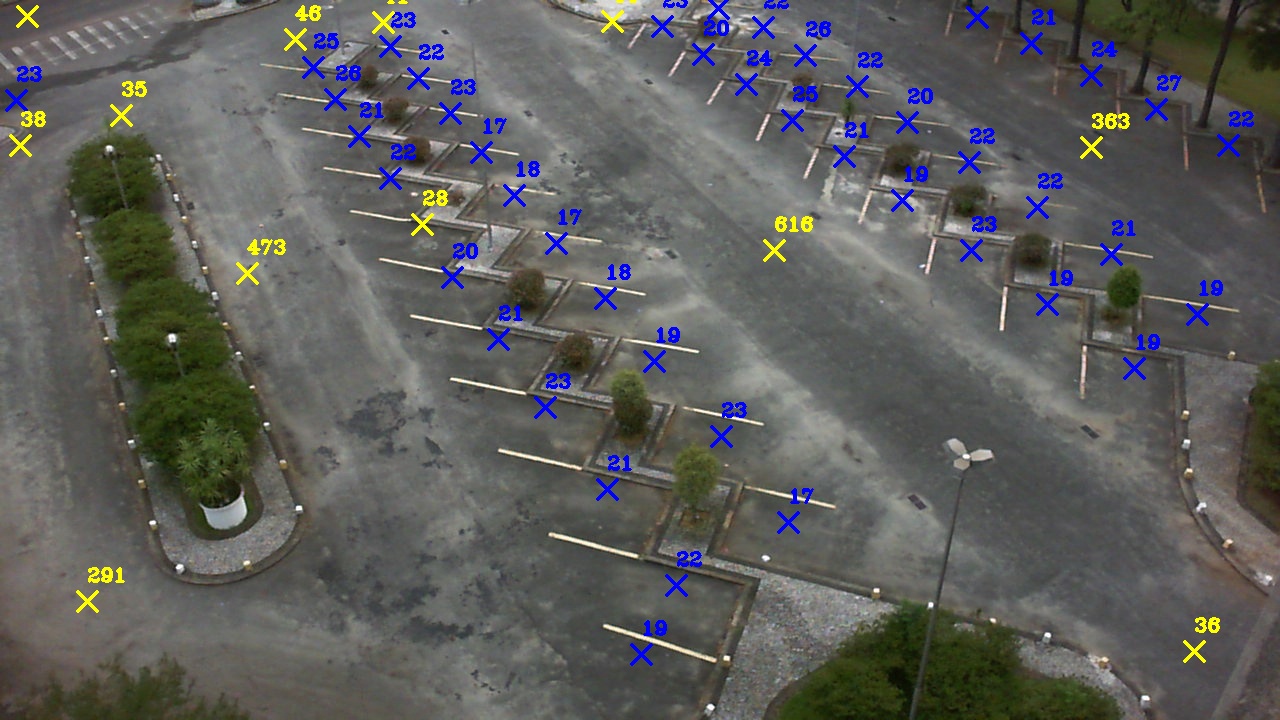}  
  \caption{Filtered cluster centers $\mathbf{m}_i$ (in blue) corresponding to the parking slots.}
  \label{fig:final_cluster_centers_pklot}
\end{subfigure}
\begin{subfigure}[t]{.45\textwidth}
  \centering
  \includegraphics[width=\linewidth]{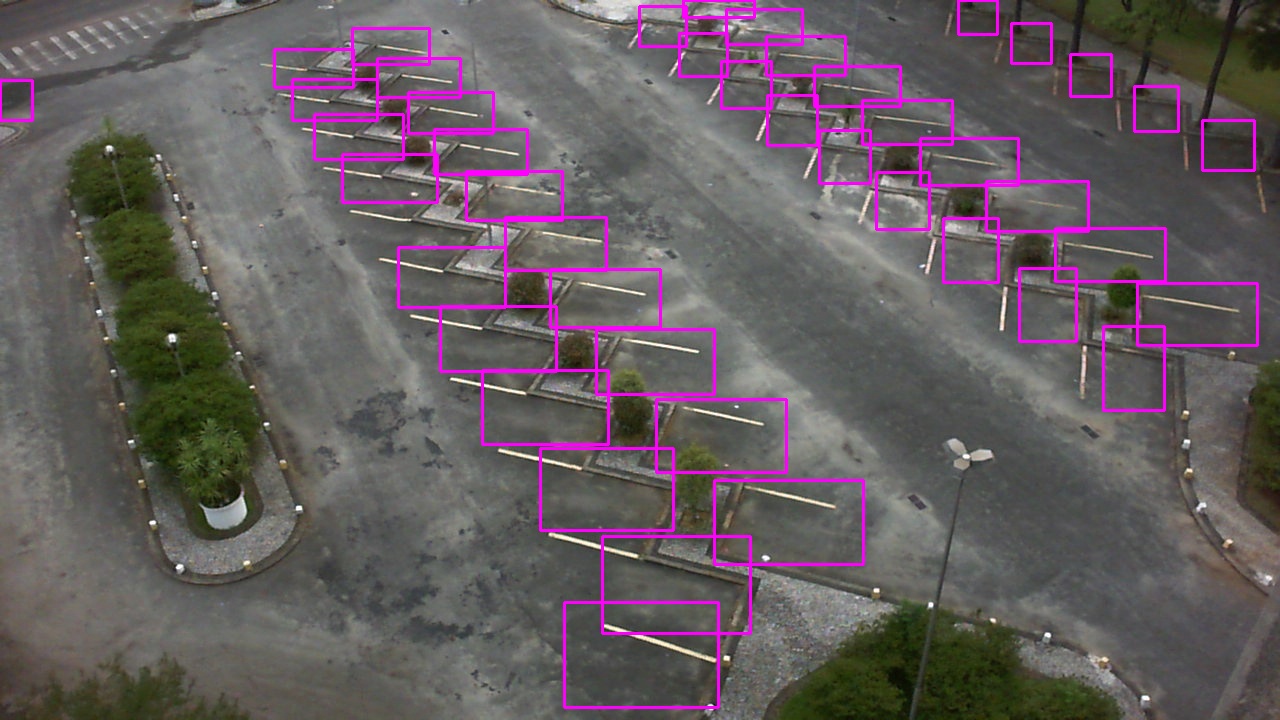}  
  \caption{Area $\mathbf{a}_i$ of each detected parking slot.\newline}
  \label{fig:final_cluster_BB_pklot}
\end{subfigure}
\par\bigskip
\begin{subfigure}[t]{.45\textwidth}
  \centering
  \includegraphics[width=\linewidth]{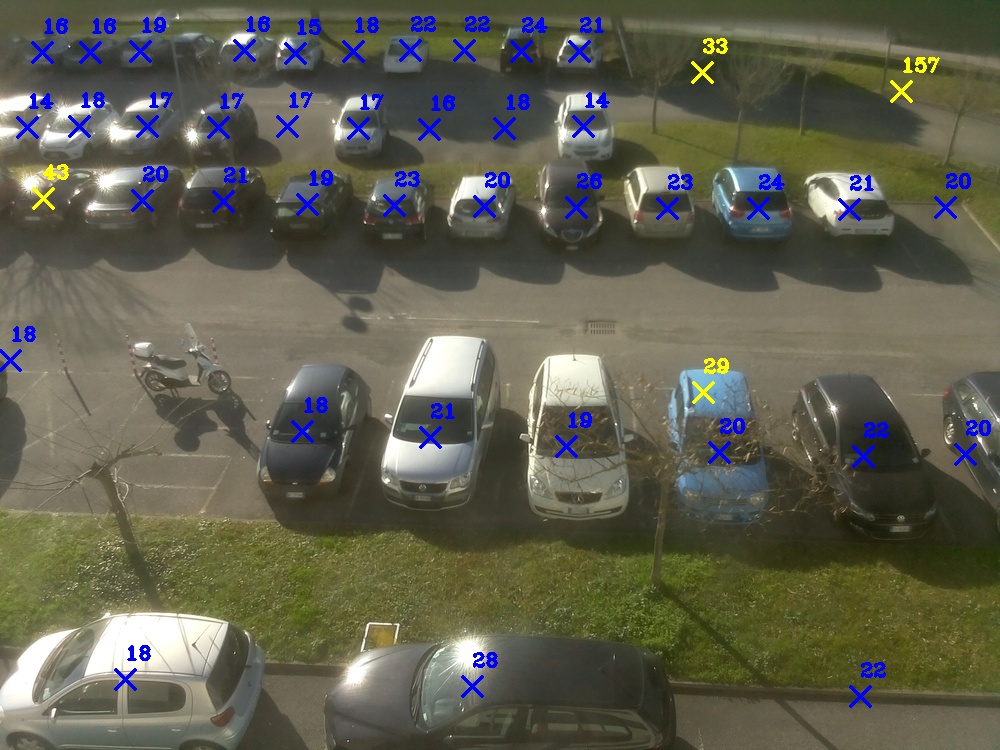}  
  \caption{Filtered cluster centers $\mathbf{m}_i$ (in blue) corresponding to the parking slots.}
  \label{fig:final_cluster_centers_cnrpark-ext}
\end{subfigure}
\begin{subfigure}[t]{.45\textwidth}
  \centering
  \includegraphics[width=\linewidth]{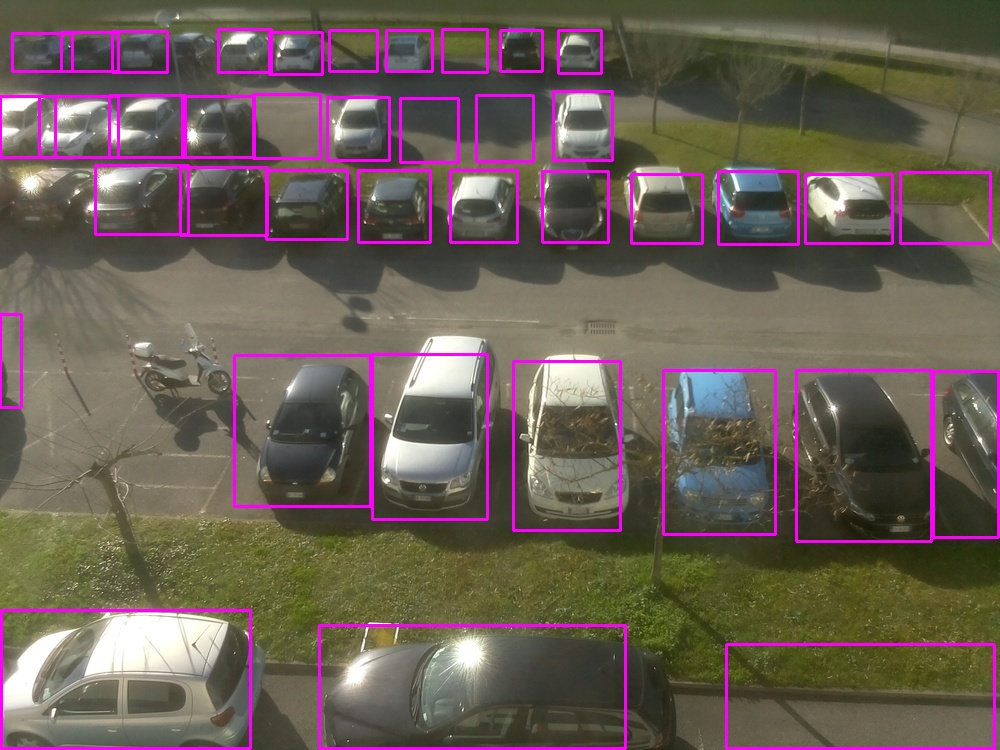}  
  \caption{Area $\mathbf{a}_i$ of each detected parking slot.}
  \label{fig:final_cluster_BB_cnrpark-ext}
\end{subfigure}
\par\bigskip
\caption{Cluster centers filtering and inverse transformation.}
\label{fig:final_result} 
\end{figure*}

The very last step in automatic parking slot detection is the transformation of cluster centers $\mathbf{m}_i$ to the original view using matrix $\mathbf{H}^{-1}$. Since the second part of the proposed APSD-OC algorithm predicts the occupancy of each parking slot, it is necessary to add area $\mathbf{a}_i$ in form of BB around each $\mathbf{m}_i$ which will be cropped and analyzed by the deep classifier. The area is defined as the mean of BBs that belong to the certain cluster. This is illustrated in Figure~\ref{fig:final_result} where $\mathbf{m}_i$ and area $\mathbf{a}_i$ are shown in original camera view for two different parking lots from PKLot and CNRPark-EXT dataset.

\subsection{Occupancy Classification Using Deep Classifier}
\label{subsec:occupancy_classification}

Obtained parking slot locations in original view are portions of an image which can be cropped and used as input to an image based parking occupancy classifier. For this purpose, a ResNet34 deep classifier pretrained on ImageNet was fine-tuned and benchmarked on various splits from the PKLot and CNRPark-EXT datasets. ResNet architecture introduced residual modules which allowed for building deeper networks without vanishing and exploding gradients. Residual modules achieve this by utilizing skip connections along which gradients flow more easily.

We followed the Amato split regarding training/testing data \cite{Amato2017}, so we can easily compare our DL classifier with current state-of-the-art. Before training, we replace the ResNet34 head with randomly initialized weights and freeze the rest of the layers. We use a learning rate finder \cite{Smith2015} to determine the base learning rate. The learning rate finder determines the optimal learning rate by starting a training with a low learning rate and doubling it for each subsequent minibatch until loss starts increasing. Our optimal learning rate is then an order of magnitude less than the learning rate when the loss starts increasing. A learning rate scheduler is also used with linear warmup on $30\%$ of batches and cosine annealing on the rest, based on the 1cycle policy \cite{Smith2017}. We also used cyclical momentum \cite{Smith2018} which we vary in the opposite direction of the learning rate. The used optimizer is Adam. Initial momentum is set to 0.95, minimum momentum is set to $0.85$ and a final momentum is set to $0.95$. We train like this for one epoch, unfreeze, then do the same over $5$ epochs with discriminative learning rates sliced from $base\_lr/200$ to $base\_lr/2$. 

\begin{figure*}[htb]
\centering
\begin{subfigure}[t]{.45\textwidth}
  \centering
  \includegraphics[width=\linewidth]{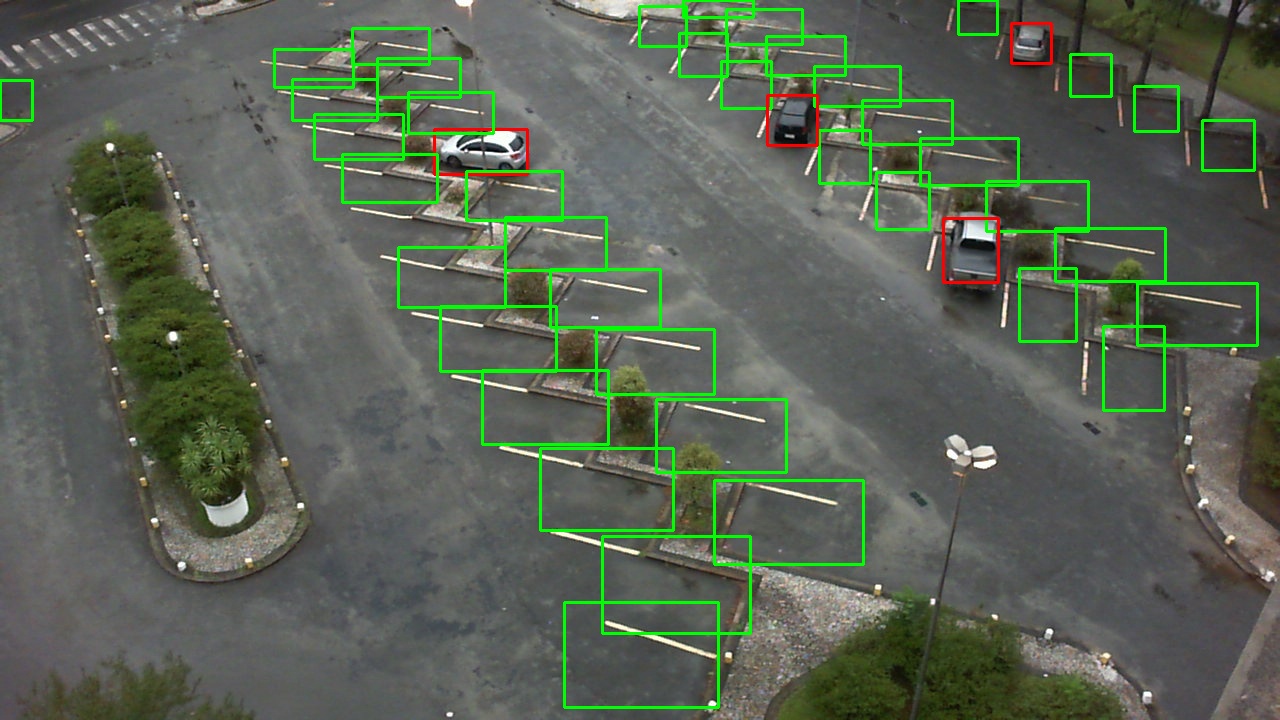}  
  \caption{Parking lot with all vehicles properly parked.}
  \label{fig:final_example_pklot_1}
\end{subfigure}
\begin{subfigure}[t]{.45\textwidth}
  \centering
  \includegraphics[width=\linewidth]{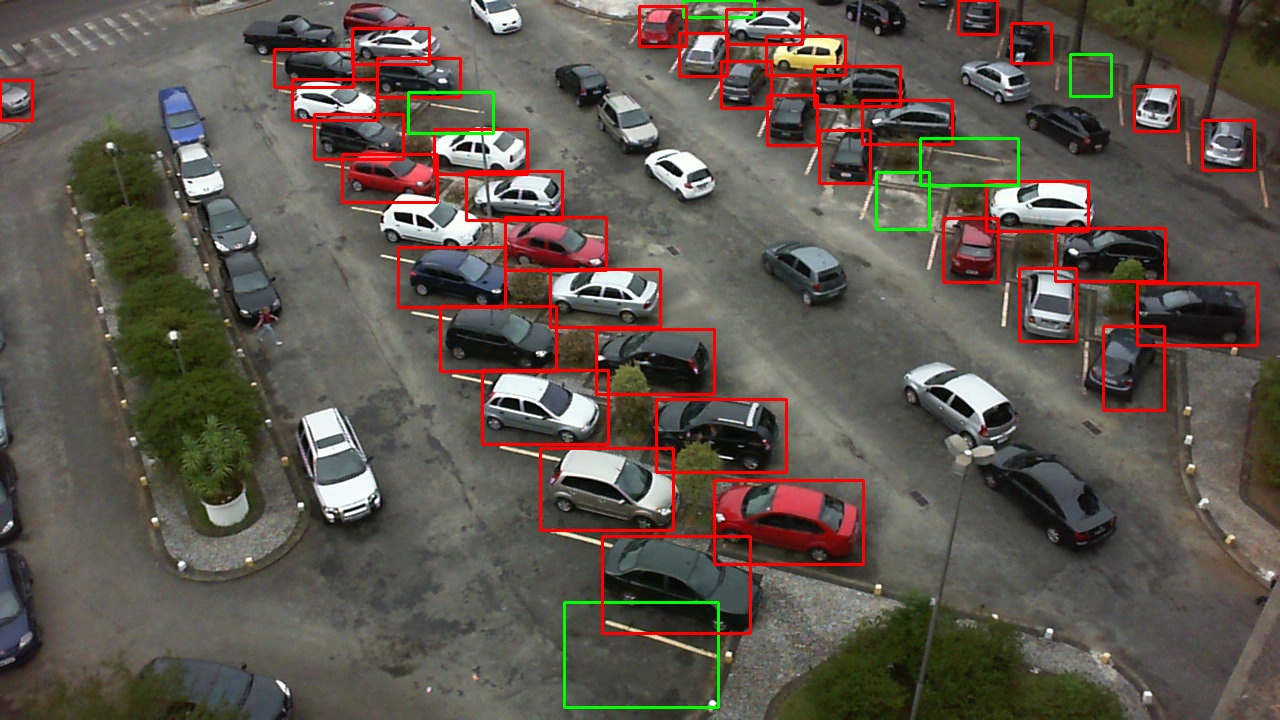}  
  \caption{Crowded parking lot with many parking violations.}
  \label{fig:final_example_pklot_2}
\end{subfigure}
\hfill
\caption{Two examples from PKLot dataset and final output of the proposed APSD-OC algorithm.}
\label{fig:final_result_examples_pklot} 
\par\bigskip
\end{figure*}

\begin{figure*}[htb]
\centering
\begin{subfigure}[t]{.45\textwidth}
  \centering
  \includegraphics[width=\linewidth]{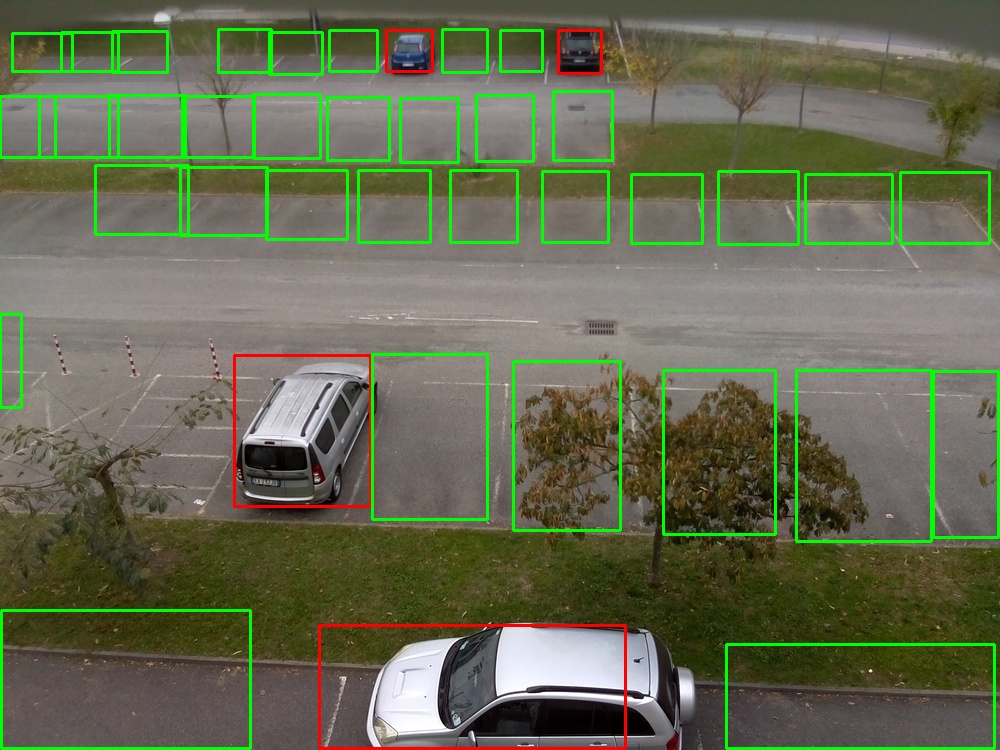}  
  \caption{Parking lot with several properly parked vehicles.\newline}
  \label{fig:final_example_cnrpark-ext_1}
\end{subfigure}
\begin{subfigure}[t]{.45\textwidth}
  \centering
  \includegraphics[width=\linewidth]{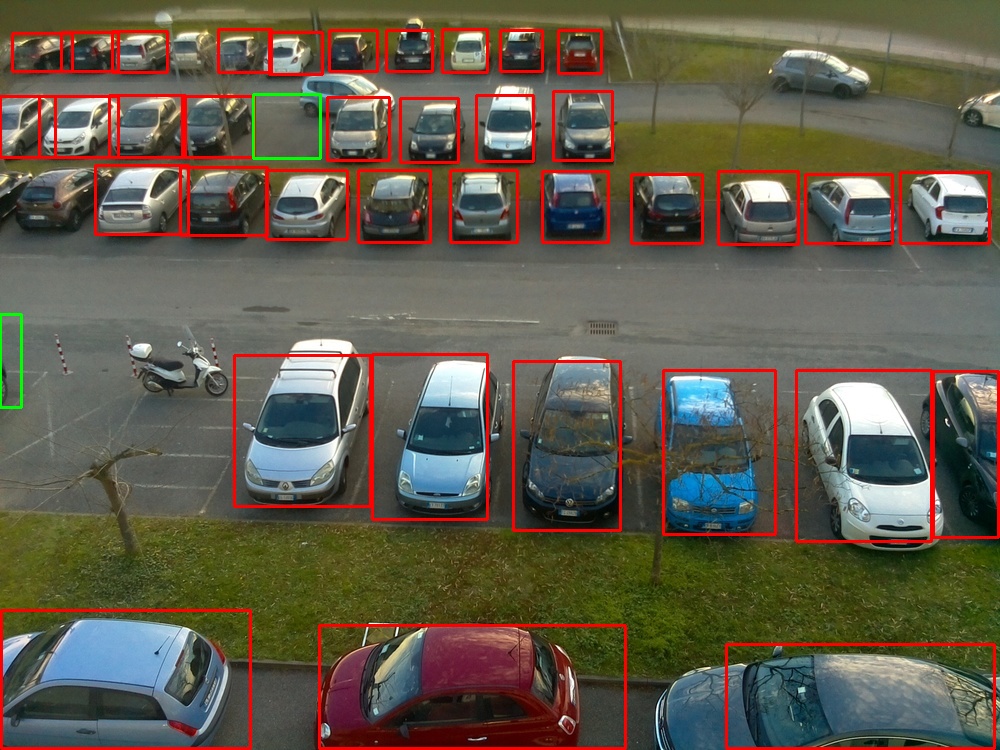}  
  \caption{Crowded parking lot with a passing vehicle and two parking violations.}
  \label{fig:final_example_cnrpark-ext_2}
\end{subfigure}
\hfill
\caption{Two examples from CNRPark-EXT dataset and final output of the proposed APSD-OC algorithm.}
\label{fig:final_result_examples_cnrpark-ext} 
\end{figure*}

Two examples of the application of the proposed APSD-OC algorithm on the parking lot from the PKLot dataset are shown in Figure~\ref{fig:final_result_examples_pklot}. It can be noticed that regular parking slots were successfully detected while the parking violations are ignored. Vacant and occupied parking slots were successfully classified by the trained deep classifier in both examples and are marked with green and red BBs, respectively. Similarly, Figure~\ref{fig:final_result_examples_cnrpark-ext} shows the example of APSD-OC algorithm application on a camera from CNRPark-EXT dataset. It can be noticed that parking slots are successfully detected although significant occlusion by the trees is present (see Figure~\ref{fig:final_example_cnrpark-ext_1}). Two parking violations and a passing vehicle are successfuly ignored in Figure~\ref{fig:final_example_cnrpark-ext_2}.

\section{Results and Discussion}
\label{sec:results_and_discussion}

\begin{figure*}[htb]
\centering
\begin{subfigure}{.32\textwidth}
  \centering
  \includegraphics[width=\linewidth]{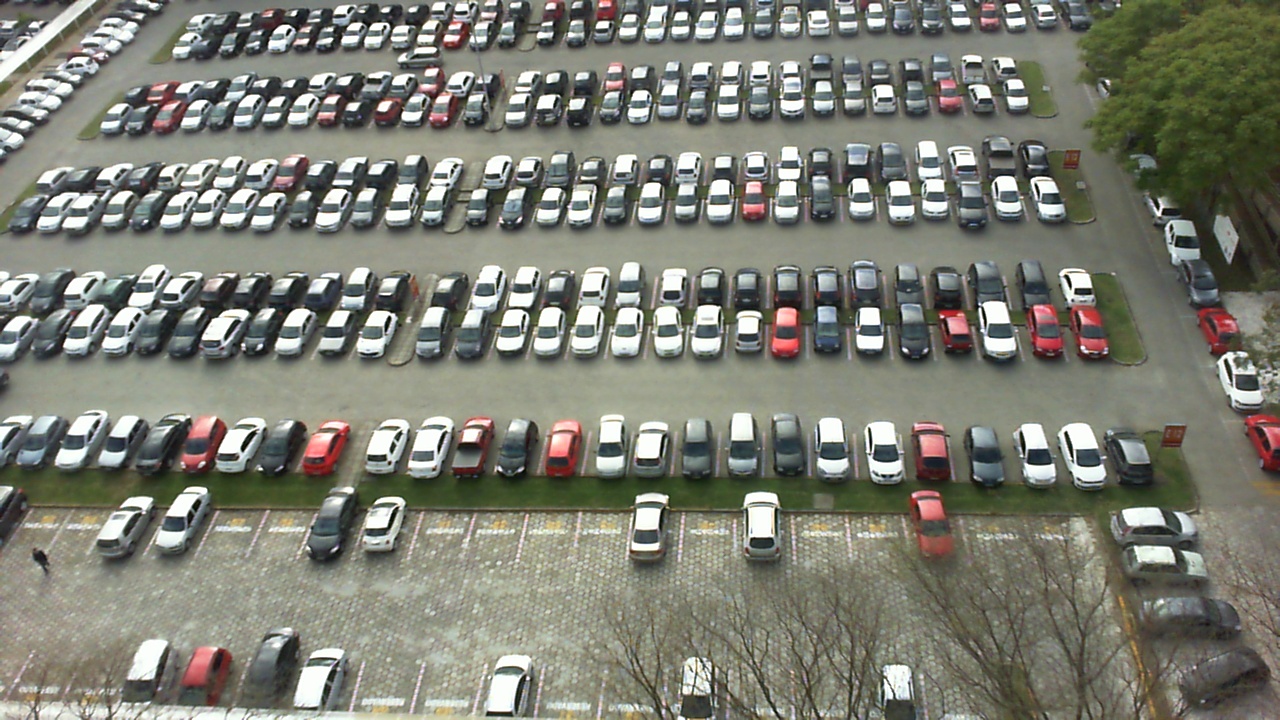}  
  \caption{PKLot PUCPR sunny}
  \label{fig:datasets:a}
\end{subfigure}
\begin{subfigure}{.32\textwidth}
  \centering
  \includegraphics[width=\linewidth]{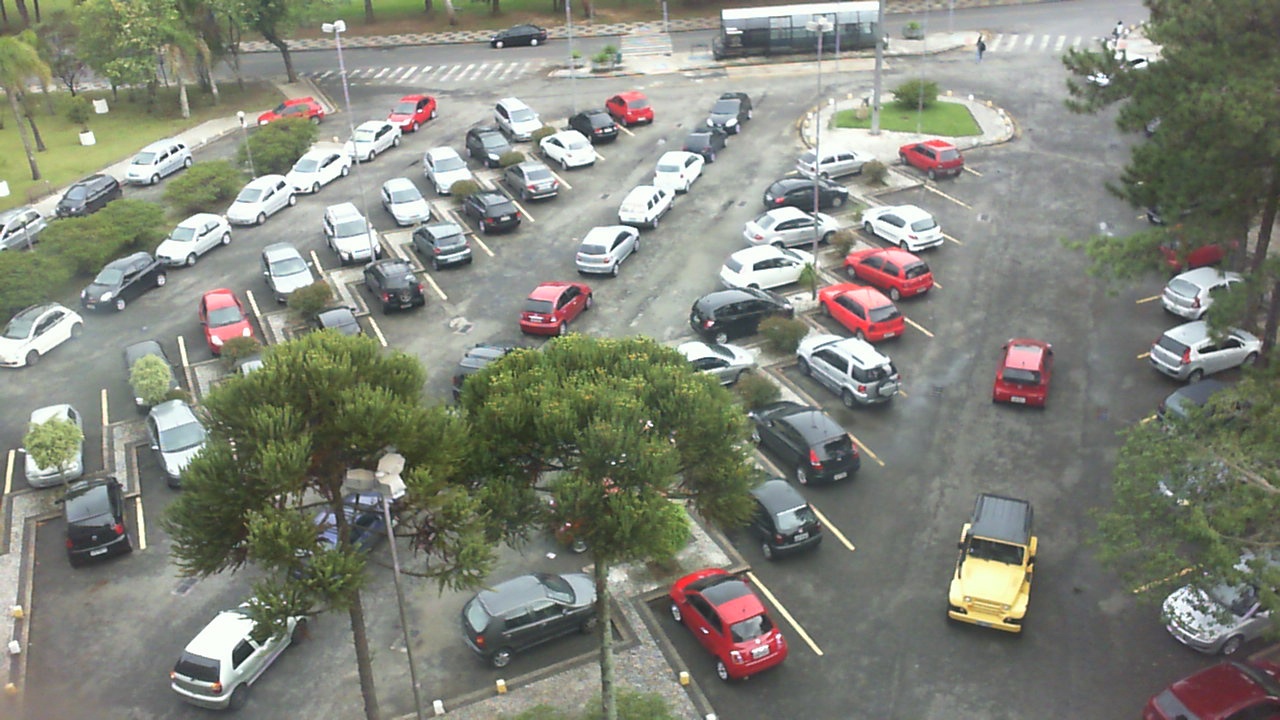}  
  \caption{PKLot UFPR04 cloudly}
  \label{fig:datasets:b}
\end{subfigure}
\begin{subfigure}{.32\textwidth}
  \centering
  \includegraphics[width=\linewidth]{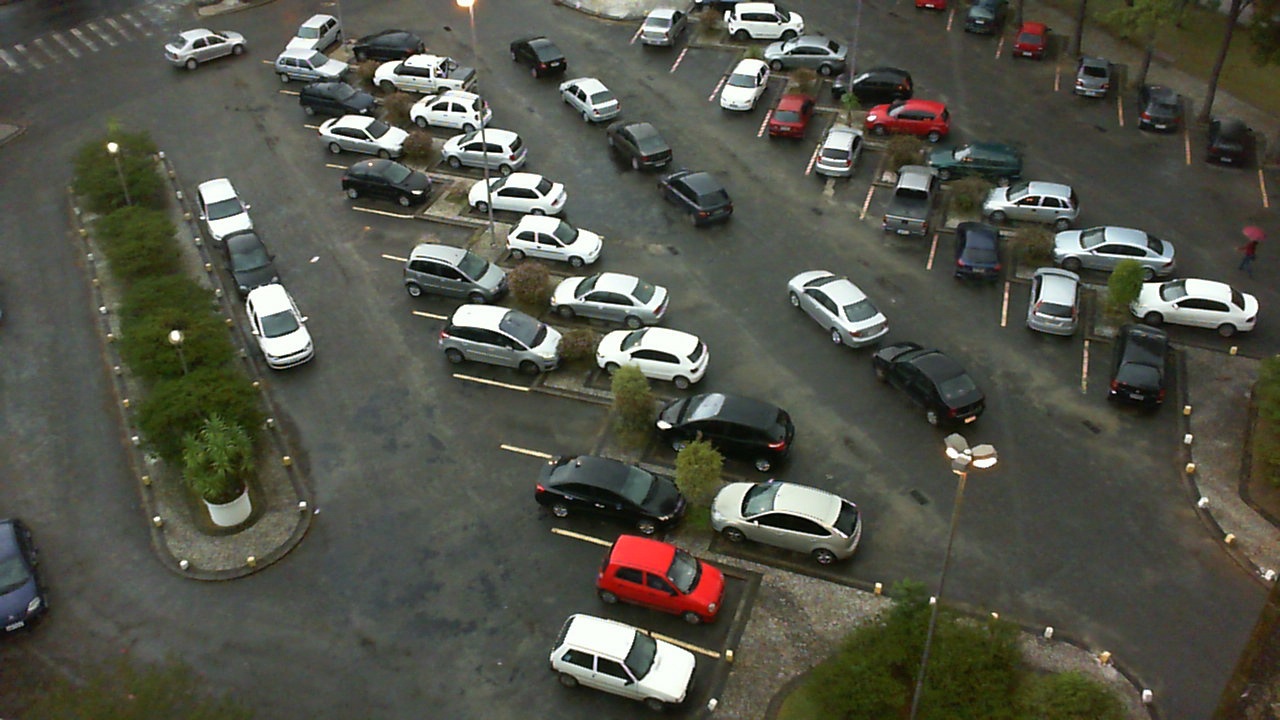}  
  \caption{PKLot UFPR05 rainy}
  \label{fig:datasets:c}
\end{subfigure}
\par\bigskip
\begin{subfigure}{.32\textwidth}
  \centering
  \includegraphics[width=\linewidth]{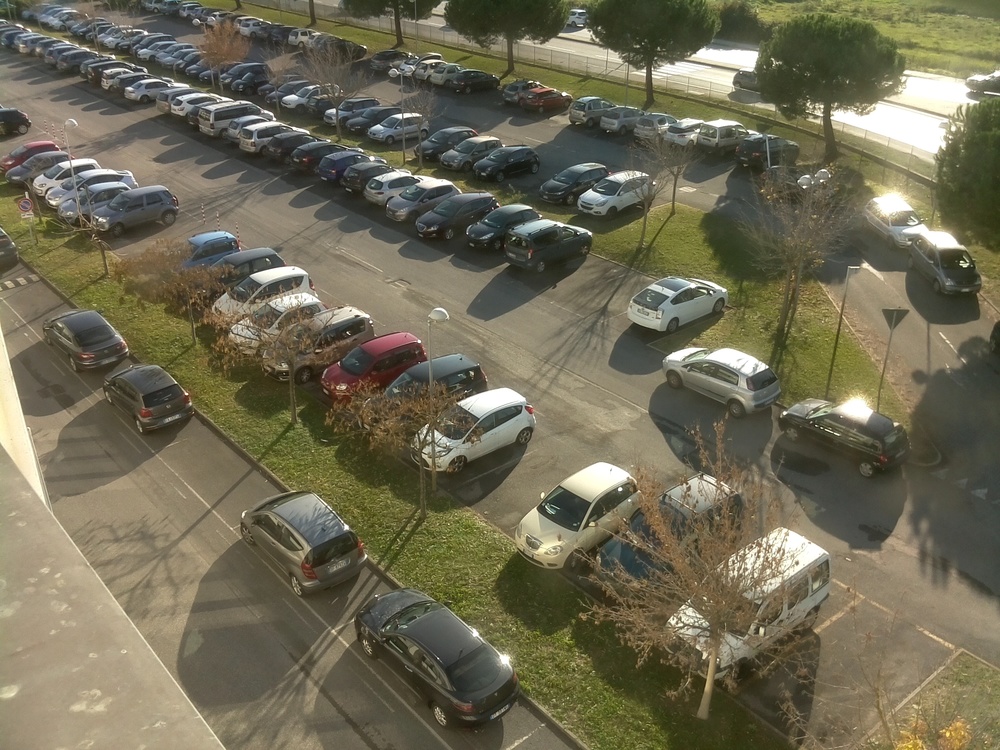}  
  \caption{CNRPark-EXT C1 sunny}
  \label{fig:datasets:d}
\end{subfigure}
\begin{subfigure}{.32\textwidth}
  \centering
  \includegraphics[width=\linewidth]{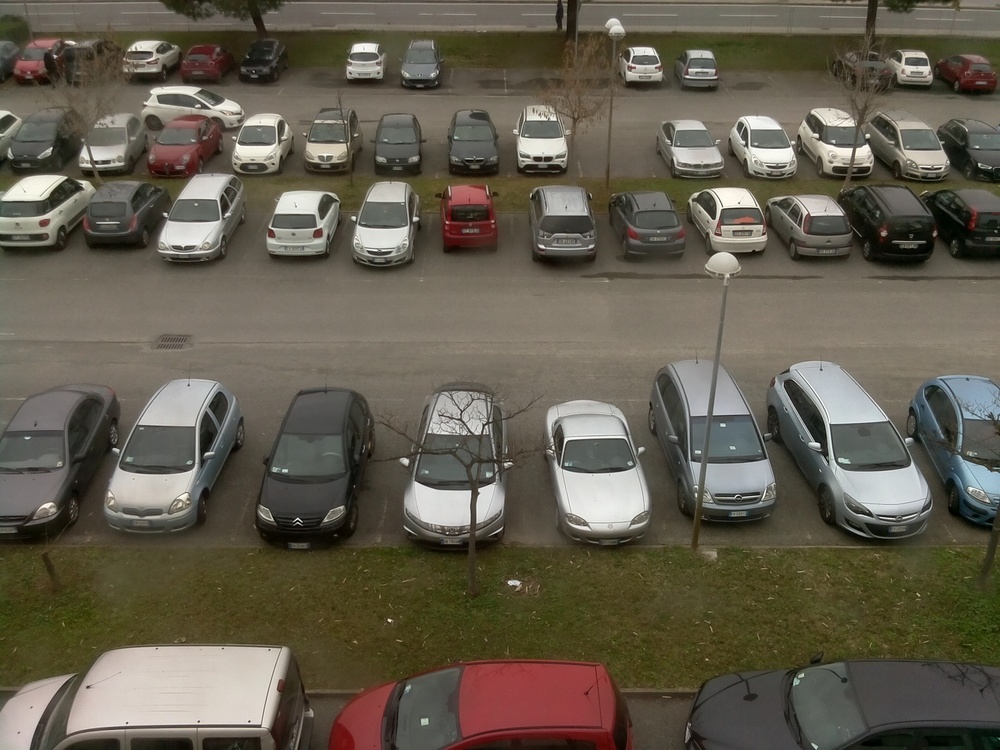}  
  \caption{CNRPark-EXT C8 overcast}
  \label{fig:datasets:e}
\end{subfigure}
\begin{subfigure}{.32\textwidth}
  \centering
  \includegraphics[width=\linewidth]{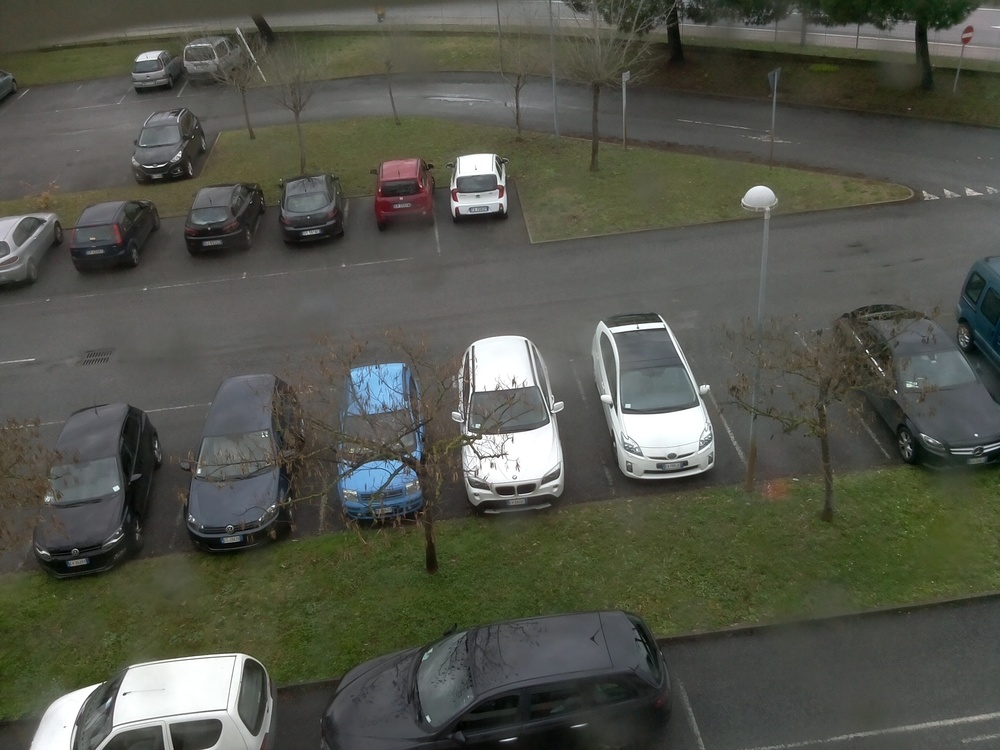}  
  \caption{CNRPark-EXT C3 rainy}
  \label{fig:datasets:f}
\end{subfigure}
\par\bigskip
\caption{Sample images from PKLot and CNRPark-EXT datasets in different weather conditions.}
\label{fig:datasets} 
\end{figure*}

Two publicly available datasets are used for APSD-OC algorithm evaluation: PKLot \cite{DeAlmeida2015} and CNRPark-EXT \cite{Amato2017}. PKLot dataset contains the images from the parking lot of the Federal University of Parana (UFPR) and the Pontifical Catholic University of Parana (PUCPR), Brazil. Images were captured every 5 minutes in time interval over 30 days. In that way, three different weather conditions are present in the dataset: sunny, cloudy and rainy. The images have 1280 x 720 pixels resolution and were stored without compression in JPEG format. Two different parking lots were captured. The images are organized into three folders with respect to the location of capturing: UFPR04, UFPR05, and PUCPR. The first two contain images of the parking lot captured from the 4th and 5th floor of the UFPR building, while the last one contains images of the parking lot captured from the 10th floor of the PUCPR administration building. Each valid parking space, i.e. parking slot which is signed with a yellow or white line is annotated with the oriented BB and can be easily extracted from the whole parking image. The CNRPark-EXT dataset is an expansion of the original CNRPark dataset \cite{Amato2017}, containing labeled images of the parking lot in the campus of the National Research Council (CNR) in Pisa, Italy. The parking lot contains a total of 164 parking slots which are captured by 9 cameras with different points of view and different perspectives from November 2015 to February 2016. In that way different weather and light conditions were captured: sunny, rainy and overcast. However, parking slots are annotated with non-rotated BBs which often do not cover precisely or entirely the parking slot area. The sample images from each dataset can be seen in Figure~\ref{fig:datasets}. The statistics regarding each dataset can be found in Table~\ref{tab:datasets}. Clearly, PKLot has significantly higher numbers of annotated parking slots. However, CNRPark-EXT has a lot of challenging images due to the occlusion of nearby objects such as lamps, trees, or other vehicles.

\begin{table*}[htb]
	\caption{PKLot and CNRPark-EXT datasets properties.}
	\label{tab:datasets}
	\footnotesize
	\begin{tabularx}{\textwidth}{XXXXXX}
	\hline
	\rowcolor{LightCyan}
	\textbf{Dataset}	& \textbf{Image resolution} &  \textbf{Sample time} &  \textbf{\# of views}	& \textbf{\# of images} & \textbf{\# of annotations}\\
        \hline
        PKLot & 1280x720 px & 5 minutes & 3 & 12,417 & 695,900 \\
        \hline
		CNRPark-EXT & 1000x750 px & 5 minutes & 9 & 4,278 & 144,965 \\
	\hline
	\end{tabularx}
\end{table*}

The evaluation of the proposed APSD-OC algorithm is performed with respect to the efficiency of the parking slot detection from the camera images and with respect to the efficiency of the parking slot occupancy classification throught the set of experiments. In the first part of experiments, images from the parking camera are fed to the algorithm which determines positions of parking slots. In the second part of experiments, cropped images of the parking slots are fed to the classifier which determines parking slot occupancy status.

\subsection{Parking Slot Detection Results}
\label{subsec:detection_results}

Although PKLot dataset and CNRPark-EXT dataset contain annotations of the regular parking slots, these cannot be used to evaluate the proposed automatic parking slot detection algorithm since some of the regular parking slots are not annotated within the dataset. This means that some false positive detection can arise during detector evaluation which are in fact regular parking spaces.

Therefore, for parking slot detection evaluation, we manually counted the number of parking slots for each parking camera in the PKLot dataset and CNRPark-EXT dataset. Only parking slots that are visible and which are marked with (white) separating lines are counted. In that way, we obtained a total of $41$, $44$, and $170$ regular parking slots for UFPR04, UFPR05 and PUCPR. In comparison, PKLot paper \cite{DeAlmeida2015} reported $28$, $40$ and $100$ parking slots for UFPR04, UFPR05, and PUCPR parking cameras. In the case of the PUCPR parking camera, we analyzed only a middle part of the camera image which contains six rows of parking slots. A similar practice was used in the PKLot dataset paper, where only part of the whole image was used for cropping of the training and testing images for classification purposes. The discarded upper part of the image corresponds to the parking area which is far from the camera and consequently, vehicles appear quite small in the image. The discarded lower part of the image corresponds to the area where vehicles are not clearly or completely visible. Regarding the CNRPark-EXT dataset, whole input camera images were used in the case of cameras C2-C8, while in the case of camera C1 and C9 part of the image was analyzed.

Standard metrics for object detections are used for efficiency of parking slot detection:
\begin{itemize}
\item	$TP$ - detection of existing parking slot
\item	$FP$ - detection of non-existing parking slot
\item	$FN$ - existing parking slot not detected
\item	$Precision = (TP)/(TP+FP)$
\item	$Recall = (TP)/(TP+FN)$
\end{itemize}

Table~\ref{tab:detection_results_PKLot} shows the obtained parking slot detection results for PKLot dataset. The available images for each camera were divided in a way that chronologically first 30\%, 50\%, 80\%, or all 100\% of camera images are used for parking slot detection. It can be noticed that FN detections are significantly dropping as the volume of the available data is increased, especially for the PUCPR parking lot. This is expected since more and more vehicles are parked inside parking slots thus satisfying $minPoints$ parameter of the DBSCAN algorithm. On the other hand, FP results are not significantly changing, proving the robustness of the proposed approach to false vehicle detections, passing vehicles, parking violations, and so on. For UFPR04 and UFPR05 parking cameras precision and recall go beyond 90\% when over 50\% of the available data is used, while in the case of PUCPR this happens at 80\% since this is a much larger parking lot. For example, the algorithm has only one FP and one FN detection for UFPR05 camera when 100\% of the data is used. This result can be also seen in Figure~\ref{fig:final_cluster_BB_pklot}. Interestingly, we noticed that UFPR04 camera is moved at a certain point in time and our assumption about fixed camera position and angle is not satisfied. Therefore, results for 80\% and 100\% of UFPR04 camera images are not reported in Table~\ref{tab:detection_results_PKLot}.

\begin{table*}[htb]
    \caption{Results of parking slot detection on PKLot dataset.}
    \label{tab:detection_results_PKLot}
    \footnotesize
    \begin{tabularx}{\textwidth}{LLLLLLLL}
    \hline
    \rowcolor{LightCyan}\textbf{Sample size} & \textbf{TP} & \textbf{FP} & \textbf{FN} & \textbf{Precision [\%]} & \textbf{Recall [\%]} &  \textbf{\# of parking slots}\\
    \hline
    \rowcolor{Gray}
    \multicolumn{7}{c}{UFPR04} \\
    \hline
    \mbox{30\%} & \mbox{36} & \mbox{5} & \mbox{5}  & \mbox{87.80}  & \mbox{87.80}  & \mbox{41}\\
    \hline
    \mbox{50\%} & \mbox{38} & \mbox{3} & \mbox{3}  & \mbox{92.68}  & \mbox{92.68}  & \mbox{41}\\ 
    \hline
    \mbox{80\%} & \multicolumn{5}{c}{Results not reported due to the camera movement} & \mbox{41}\\ 
    \hline
    \mbox{100\%} & \multicolumn{5}{c}{Results not reported due to the camera movement} & \mbox{41}\\ 
	\hline
	\rowcolor{Gray}
	\multicolumn{7}{c}{UFPR05} \\
	\hline
	\mbox{30\%} & \mbox{36} & \mbox{8} & \mbox{8}  & \mbox{81.82}  & \mbox{81.82}  & \mbox{44}\\
	\hline
    \mbox{50\%} & \mbox{40} & \mbox{3} & \mbox{4}  & \mbox{93.02}  & \mbox{90.91}  & \mbox{44}\\ 
    \hline
    \mbox{80\%} & \mbox{42} & \mbox{2} & \mbox{2}  & \mbox{95.45}  & \mbox{95.45}  & \mbox{44}\\ 
    \hline
    \mbox{100\%} & \mbox{43} & \mbox{1} & \mbox{1}  & \mbox{97.73} & \mbox{97.73}  & \mbox{44}\\ 
	\hline
	\rowcolor{Gray}
	\multicolumn{7}{c}{PUCPR} \\
	\hline
	\mbox{30\%} & \mbox{142} & \mbox{11} & \mbox{28}  & \mbox{92.81}  & \mbox{83.53}  & \mbox{170}\\
	\hline
    \mbox{50\%} & \mbox{147} & \mbox{10} & \mbox{23}  & \mbox{96.63}  & \mbox{86.47}  & \mbox{170}\\
    \hline
    \mbox{80\%} & \mbox{155} & \mbox{9} & \mbox{15}  & \mbox{94.51}  & \mbox{91.18}  & \mbox{170}\\
    \hline
    \mbox{100\%} & \mbox{158} & \mbox{7} & \mbox{12}  & \mbox{95.76}  & \mbox{92.94}  & \mbox{170}\\
	\hline
    \end{tabularx}
\end{table*}

Table~\ref{tab:detection_results_CNRPark-EXT} shows the obtained parking slot detection result for CNRPark-EXT dataset. Hereby, the same pattern can be observed like in PKLot dataset results - more data mostly ensure higher precision and recall in parking slot detection. Interestingly, for the number of cameras (C2, C4, C5, C6, C7, C8, C9) 100\% precision is obtained since in these camera images there are practically no parking violations so each detection corresponds to the actual parking slot. However, for most of the cameras obtained recall is somewhat lower than for the PKLot dataset which can be attributed to the occlusion of parking slots with nearby objects like trees. In such parking slots vehicles are often not properly detected in first algorithm step.

\begin{table*}[p]
    \caption{Results of parking slot detection on CNRPark-EXT dataset.}
    \label{tab:detection_results_CNRPark-EXT}
	\scriptsize
    \begin{tabularx}{\textwidth}{LLLLLLLL}
    \hline
    \rowcolor{LightCyan}\textbf{Sample size} & \textbf{TP} & \textbf{FP} & \textbf{FN} & \textbf{Precision [\%]} & \textbf{Recall [\%]} &  \textbf{\# of parking slots}\\
   
    \hline
    \rowcolor{Gray}
    \multicolumn{7}{c}{Camera 1} \\
    \hline
    \mbox{30\%} & 29	& 6		& 7		& 82.86		& 80.56		& 36 \\
    \hline
    \mbox{50\%} & 31	& 4		& 5		& 88.57  	& 86.11		& 36 \\
    \hline
    \mbox{80\%}	& 26	& 4		& 10	& 86.67		& 72.22 	& 36 \\
    \hline
    \mbox{100\%} & 30	& 5		& 6		& 85.71 	& 83.33		& 36 \\
    
	\hline
	\rowcolor{Gray}
    \multicolumn{7}{c}{Camera 2} \\
    \hline
    \mbox{30\%} & 9		& 1		& 2		& 90.00		& 81.82		& 11 \\
    \hline
    \mbox{50\%} & 8		& 3		& 3		& 72.73		& 72.73		& 11 \\
    \hline
    \mbox{80\%} & 10	& 0		& 1		& 100.00  	& 90.91		& 11 \\
    \hline
    \mbox{100\%} & 11	& 0		& 0		& 100.00  	& 100.00	& 11 \\
    
    \hline	
    \rowcolor{Gray}
    \multicolumn{7}{c}{Camera 3} \\
    \hline
    \mbox{30\%} & 22	& 4		& 4		& 84.61		& 84.61		& 26 \\
    \hline
    \mbox{50\%} & 24	& 2		& 2		& 92.31		& 92.31		& 26 \\
    \hline
    \mbox{80\%} & 23	& 3		& 3		& 88.46  	& 88.46		& 26 \\
    \hline
    \mbox{100\%} & 25	& 1		& 1		& 96.15		& 96.15		& 26 \\
    
    \hline
    \rowcolor{Gray}	
    \multicolumn{7}{c}{Camera 4} \\
    \hline
    \mbox{30\%} & 36	& 3		& 5		& 92.31		& 87.80    	& 41 \\
   	\hline
    \mbox{50\%} & 38	& 0		& 3		& 100.00	& 92.68		& 41 \\
    \hline
    \mbox{80\%} & 38 	& 1		& 3		& 97.44		& 92.68		& 41 \\
    \hline
    \mbox{100\%} & 39	& 0		& 2		& 100.00    & 95.12		& 41 \\
    
    \hline
    \rowcolor{Gray}
    \multicolumn{7}{c}{Camera 5} \\
    \hline
    \mbox{30\%} & 46	& 0		& 7		& 100.00	& 86.79		& 53 \\
    \hline
    \mbox{50\%} & 46	& 0		& 7		& 100.00	& 86.79		& 53 \\	
    \hline
    \mbox{80\%} & 49	& 0		& 4		& 100.00	& 92.45		& 53 \\
    \hline
    \mbox{100\%} & 49  	& 0		& 4		& 100.00	& 92.45		& 53 \\
    
    \hline
    \rowcolor{Gray}
    \multicolumn{7}{c}{Camera 6} \\
    \hline
    \mbox{30\%} & 44	& 1		& 9		& 97.78		& 83.02		& 53 \\
    \hline
    \mbox{50\%} & 48	& 1		& 5		& 97.96		& 90.57		& 53 \\
    \hline
    \mbox{80\%} & 48	& 0		& 5		& 100.00	& 90.57		& 53 \\
    \hline
    \mbox{100\%} & 48	& 0		& 5		& 100.00	& 90.57		& 53 \\
    
    \hline
    \rowcolor{Gray}
    \multicolumn{7}{c}{Camera 7} \\
    \hline
    \mbox{30\%} & 48	& 0		& 9		& 100.00	& 84.21		& 57 \\
    \hline
    \mbox{50\%} & 49	& 0 	& 8		& 100.00	& 85.96		& 57 \\
    \hline
    \mbox{80\%} & 52	& 0		& 5		& 100.00	& 91.23		& 57 \\
    \hline
    \mbox{100\%} & 52	& 0		& 5		& 100.00	& 91.23		& 57 \\
    
    \hline
    \rowcolor{Gray}
    \multicolumn{7}{c}{Camera 8} \\
    \hline
    \mbox{30\%} & 52	& 0		& 4		& 100.00	& 92.86		& 56 \\
    \hline
    \mbox{50\%} & 49  	& 0		& 7  	& 100.00	& 87.50		& 56 \\
    \hline
    \mbox{80\%} & 51  	& 0		& 5  	& 100.00	& 91.07		& 56 \\
    \hline
    \mbox{100\%} & 51  	& 0		& 5  	& 100.00	& 91.07		& 56 \\
    
    \hline
    \rowcolor{Gray}
    \multicolumn{7}{c}{Camera 9} \\
    \hline
    \mbox{30\%} & 29	& 0		& 3		& 100.00	& 90.63		& 32 \\
    \hline
    \mbox{50\%} & 31	& 0		& 1		& 100.00	& 96.88		& 32 \\
    \hline
    \mbox{80\%} & 30	& 0		& 2		& 100.00	& 93.75		& 32 \\
    \hline
    \mbox{100\%} & 29	& 0		& 3   	& 100.00   	& 90.63		& 32 \\
    
	\hline
    \end{tabularx}
\end{table*}

\subsection{Parking Slot Occupancy Classification Results}
\label{subsec:classification_results}

For the parking slot occupancy classification standard metrics are used:
\begin{itemize}
\item	$TP$ - occupied parking slot classified as occupied,
\item	$TN$ - vacant parking slot classified as vacant,
\item	$FP$ - vacant parking slot classified as occupied,
\item	$FN$ - occupied parking slot classified as vacant,
\item	$Accuracy = (TP+TN)/(TP+TN+FP+FN)$,
\item	$AUC$ - area under the Receiver Operating Characteristic (ROC). 
\end{itemize}

To make a fair comparison with recent approaches, we followed guidelines from PKLot dataset~\cite{DeAlmeida2015} and considered 50\% of the images available in the subsets UFPR04, UFPR05, and PUCPR for training and 50\% for testing. We performed single parking lot training and multiple parking lot testing since we are mostly interested in the generalization power of classifiers. This results in three different classifiers where each classifier is evaluated on three different test datasets. The number of images containing occupied or vacant parking slots can be found in Table~\ref{tab:PKLot_division}. It can be noticed that the PUCPR parking lot has a higher number of images in comparison with UFPR04 and UFPR05 since the corresponding camera covers a significantly higher number of parking slots. Also, subsets are imbalanced, vacant slots being majority class for UFPR04 and PUCPR parking lots, and occupied slots being majority class for UFPR05 parking lot. However, this imbalance is not significant and does not require any special approach to classifier training and testing.

\begin{table*}[htb]

    \caption{Summary of created PKLot subsets.}
    \label{tab:PKLot_division}
    \footnotesize
    \begin{tabularx}{\textwidth}{XXXXXX}
	\hline
	\rowcolor{LightCyan}
	
	& & \multicolumn{3}{c}{\textbf{\# of parking slots}} \\
	
	\rowcolor{LightCyan}
	\multirow{-2}{*}{\textbf{Parking lot}} & \multirow{-2}{*}{\textbf{Subset}} & \textbf{Occupied} & \textbf{Vacant} & \textbf{Total}	\\ 
	\hline
	\multirow{3}{*}{UFPR04}   	& Train		& 23,050		& 29,871    	& 52,921	 	\\
    	                    	& Test		& 23,075		& 29,847		& 52,922	 	\\ \cline{3-5}
    	                    	& Total		& 46,125		& 59,718		& 105,843		\\
    	                    	
	\hline
   	\multirow{3}{*}{UFPR05}   	& Train   	& 48,967     	& 33,925     	& 82,892 		\\
    	                    	& Test   	& 48,459     	& 34,434		& 82,893	 	\\ \cline{3-5}
    	                    	& Total		& 97,426		& 68,359		& 165,785		\\
    \hline
    \multirow{3}{*}{PUCPR}   	& Train   	& 96,736     	& 115,375    	& 212,111		\\
    	                    	& Test   	& 97,493     	& 114,619    	& 212,112		\\ \cline{3-5}
    	                    	& Total		& 194,229		& 229,994		& 424,223		\\
    \hline
    \end{tabularx}
\end{table*}

The comparison regarding classification accuracy is made with CarNet~\cite{Nurullayev2019} and mAlexNet~\cite{Amato2017}. The obtained results are presented in Table~\ref{tab:classification_accuracy_results_PKLot} for different combinations of training and testing subsets. The best result is shown in bold for each considered case. The obtained results show that our proposed deep classifier obtains the best result in seven out of nine possible subset combinations. This strongly indicates that the proposed approach to parking occupancy classifier learning results in high generalization power. This is especially pronounced when learning on the PUCPR subset and testing on the UFPR04 subset where significantly different camera perspectives are present. In this case, our approach has 4\% higher accuracy than state-of-the-art approach CarNet~\cite{Nurullayev2019}.

\begin{table*}[htb]
	\caption{Accuracy of parking slots classification on PKLot dataset.}
    \label{tab:classification_accuracy_results_PKLot}
    \footnotesize
    \begin{tabularx}{\textwidth}{XXXXX}
	\hline
	\rowcolor{LightCyan}
	& & \multicolumn{3}{c}{\textbf{Testing accuracy}} \\
	\rowcolor{LightCyan}
	\multirow{-2}{*}{\textbf{Method}} & \multirow{-2}{*}{\textbf{Training subset}} & \mbox{\textbf{UFPR04}} & \mbox{\textbf{UFPR05}} & \mbox{\textbf{PUCPR}}	\\ 
	\hline
	\multirow{3}{*}{Ours}   & UFPR04   & \textbf{99.98\%}      & 95.47\%     & \textbf{99.19\%} 	\\
    	                    & UFPR05   & \textbf{95.29\%}      & \textbf{99.92\%}     & 98.08\%		\\
            	            & PUCPR    & \textbf{98.62\%}      & \textbf{98.60\%}     & \textbf{99.93\%}	\\ 
	\hline
	\multirow{3}{*}{CarNet}	& UFPR04   & 95.60\%	& \textbf{97.60\%}	& 98.30\%	\\
                        							& UFPR05   & 95.20\%	& 97.50\%	& \textbf{98.40\%}   \\
                        							& PUCPR    & 94.40\%	& 97.70\%   & 98.80\%    \\      
    \hline
	\multirow{3}{*}{mAlexNet} 	& UFPR04   & 99.54\%	& 93.29\%	& 98.27\%	\\
                        							& UFPR05   & 93.69\%	& 99.49\%	& 92.72\%   \\
                        							& PUCPR    & 98.03\%	& 96.00\%   & 99.90\%    \\      
    \hline   
    \end{tabularx}
\end{table*}

The proposed classifier efficiency on the PKLot dataset in terms of AUC is reported in Table~\ref{tab:classification_AUC_results_PKLot}. The comparison is performed with CarNet~\cite{Nurullayev2019}, mAlexNet~\cite{Amato2017} and the approach based on integral channel features proposed in \cite{Ahrnbom2016}. Again, our proposed approach achieves the best result in seven out of nine possible subset combinations and obtains AUC greater than $0.99$ for each subset combination thus confirming its high generalization abilities.

\begin{table*}[htb]
    \caption{Results of parking slot occupancy classification on PKLot dataset.}
    \label{tab:classification_AUC_results_PKLot}
    \scriptsize
    \begin{tabularx}{\textwidth}{LLLL}
    \hline
    \rowcolor{LightCyan}\textbf{Method} & \textbf{Training subset} & \textbf{Testing subset} & \textbf{AUC} \\
    \hline
    \mbox{Ours} 	& \mbox{UFPR04} & \mbox{UFPR04} & \mbox{\textbf{0.9999}}		\\
    \mbox{CarNet} 	& \mbox{UFPR04} & \mbox{UFPR04} & \mbox{0.9790}	\\
    \mbox{ICF+LR}	& \mbox{UFPR04} & \mbox{UFPR04} & \mbox{0.9994}		\\
    \mbox{ICF+SVM}  & \mbox{UFPR04} & \mbox{UFPR04} & \mbox{0.9996}	\\
    \mbox{PKLot}	& \mbox{UFPR04} & \mbox{UFPR04} & \mbox{\textbf{0.9999}}	\\
    \hline 
    \mbox{Ours} 	& \mbox{UFPR04} & \mbox{UFPR05} & \mbox{\textbf{0.9989}}		\\
    \mbox{CarNet} 	& \mbox{UFPR04} & \mbox{UFPR05} & \mbox{0.9935}	\\
    \mbox{ICF+LR}	& \mbox{UFPR04} & \mbox{UFPR05} & \mbox{0.9928}		\\
    \mbox{ICF+SVM}  & \mbox{UFPR04} & \mbox{UFPR05} & \mbox{0.9772}	\\
    \mbox{PKLot}	& \mbox{UFPR04} & \mbox{UFPR05} & \mbox{0.9595}	\\
    \hline    
    \mbox{Ours} 	& \mbox{UFPR04} & \mbox{PUCPR} & \mbox{\textbf{0.9995}}		\\
    \mbox{CarNet}	& \mbox{UFPR04} & \mbox{PUCPR} & \mbox{0.9982}	\\
    \mbox{ICF+LR}	& \mbox{UFPR04} & \mbox{PUCPR} & \mbox{0.9881}		\\
    \mbox{ICF+SVM}	& \mbox{UFPR04} & \mbox{PUCPR} & \mbox{0.9569}	\\
    \mbox{PKLot}	& \mbox{UFPR04} & \mbox{PUCPR} & \mbox{0.9713}	\\
    \hline
    
    \mbox{Ours} 	& \mbox{UFPR05} & \mbox{UFPR04} & \mbox{0.9943}		\\
    \mbox{CarNet}	& \mbox{UFPR05} & \mbox{UFPR04} & \mbox{0.9796}	\\
    \mbox{ICF+LR}	& \mbox{UFPR05} & \mbox{UFPR04} & \mbox{\textbf{0.9963}}	\\
    \mbox{ICF+SVM}	& \mbox{UFPR05} & \mbox{UFPR04} & \mbox{0.9943}	\\
    \mbox{PKLot}	& \mbox{UFPR05} & \mbox{UFPR04} & \mbox{0.9533}	\\
    \hline 
    \mbox{Ours} 	& \mbox{UFPR05} & \mbox{UFPR05} & \mbox{\textbf{0.9999}}		\\
    \mbox{CarNet} 	& \mbox{UFPR05} & \mbox{UFPR05} & \mbox{0.9989}	\\
    \mbox{ICF+LR}	& \mbox{UFPR05} & \mbox{UFPR05} & \mbox{0.9987}		\\
    \mbox{ICF+SVM}	& \mbox{UFPR05} & \mbox{UFPR05} & \mbox{0.9988}	\\
    \mbox{PKLot}	& \mbox{UFPR05} & \mbox{UFPR05} & \mbox{0.9995}	\\
    \hline    
    \mbox{Ours} 	& \mbox{UFPR05} & \mbox{PUCPR} & \mbox{\textbf{0.9978}}		\\
    \mbox{CarNet}	& \mbox{UFPR05} & \mbox{PUCPR} & \mbox{0.9791}	\\
    \mbox{ICF+LR}	& \mbox{UFPR05} & \mbox{PUCPR} & \mbox{0.9779}		\\
    \mbox{ICF+SVM}	& \mbox{UFPR05} & \mbox{PUCPR} & \mbox{0.9405}	\\
    \mbox{PKLot}	& \mbox{UFPR05} & \mbox{PUCPR} & \mbox{0.9761}	\\
    \hline
    
    \mbox{Ours} 	& \mbox{PUCPR} & \mbox{UFPR04} & \mbox{\textbf{0.9985}}		\\
    \mbox{CarNet} 	& \mbox{PUCPR} & \mbox{UFPR04} & \mbox{0.9845}	\\
    \mbox{ICF+LR} 	& \mbox{PUCPR} & \mbox{UFPR04} & \mbox{0.9829}		\\
    \mbox{ICF+SVM} 	& \mbox{PUCPR} & \mbox{UFPR04} & \mbox{0.9843}	\\
    \mbox{PKLot}	& \mbox{PUCPR} & \mbox{UFPR04} & \mbox{0.9589}	\\
    \hline 
    \mbox{Ours}		& \mbox{PUCPR} & \mbox{UFPR05} & \mbox{\textbf{0.9981}}		\\
    \mbox{CarNet}	& \mbox{PUCPR} & \mbox{UFPR05} & \mbox{0.9938}	\\
    \mbox{ICF+LR}	& \mbox{PUCPR} & \mbox{UFPR05} & \mbox{0.9457}		\\
    \mbox{ICF+SVM}	& \mbox{PUCPR} & \mbox{UFPR05} & \mbox{0.9401}	\\
    \mbox{PKLot}	& \mbox{PUCPR} & \mbox{UFPR05} & \mbox{0.9152}	\\
    \hline   
    \mbox{Ours}		& \mbox{PUCPR} & \mbox{PUCPR} & \mbox{0.9998}		\\
    \mbox{CarNet}	& \mbox{PUCPR} & \mbox{PUCPR} & \mbox{0.9986}	\\
    \mbox{ICF+LR}	& \mbox{PUCPR} & \mbox{PUCPR} & \mbox{0.9994}		\\
    \mbox{ICF+SVM}	& \mbox{PUCPR} & \mbox{PUCPR} & \mbox{0.9994}	\\
    \mbox{PKLot}	& \mbox{PUCPR} & \mbox{PUCPR} & \mbox{\textbf{0.9999}}	\\
    \hline
    \end{tabularx}
\end{table*}

In the case of the CNRPark-EXT dataset, we also followed the dataset split proposed in \cite{Amato2017} so we can directly compare our classifier with mAlexNet, AlexNet and partially with \cite{Nurullayev2019}. According to Table~\ref{tab:classification_AUC_results_CNRPark-EXT}, our proposed classifier outperforms mAlexNet and AlexNet for each train/test subset combination by a large margin. For example, when training our classifier only on CNRPark dataset, which contains images from only two cameras and images captured during sunny days, it obtains significantly higher accuracy (more than 4\%) than AlexNet when testing on CNRPark-EXT TEST dataset which contains 9 cameras and different weather conditions. When the training subset contains CNRPark dataset and images from CNRPark-EXT train C1-C8, then accuracy on CNRPark-EXT TEST dataset goes over 99\% indicating that very high classifier accuracy can be obtained if the training subset contains different viewpoints and weather conditions. The proposed classifier also outperforms state-of-the-art CarNet on CNRPark-EXT dataset as shown in Table~\ref{tab:classification_AUC_results_CNRPark-EXT}.

More detailed experiments were conducted related to viewpoint changes and weather conditions like in \cite{Amato2017}. The classifier was trained on images taken with a single camera (C1 and C8) and tested against images of all other cameras in the CNRPark-EXT dataset. The obtained results regarding viewpoint generalization of the built classifier are reported in Table~\ref{tab:classification_AUC_results_CNRPark-EXT_viewpoint}. In almost all cases proposed classifier outperforms mAleXNet, achieving accuracy over 95\% in all cases except in case of training on images taken by camera C8 and testing on images taken by camera C1 since these are two cameras with significantly different viewpoints (C1 is a side view of a parking lot and C8 is a front view of a parking lot). However, the difference in the obtained accuracy of the proposed classifier on the same test camera does not differ more than 1\% when training on C1 or C8 which is not true for mAlexNet. It can be concluded that the proposed classifier has better viewpoint robustness.

Similar behavior can be observed when training the classifier on images captured during particular weather condition (sunny, rainy, or overcast) and testing against images captured during remaining weather conditions. The obtained results regarding weather generalization are shown in Table~\ref{tab:classification_AUC_results_CNRPark-EXT_weather}. It can be noticed that the proposed classsifier significantly outperforms mAlexNet and obtains accuracy over 98\% in all cases. Certainly, the proposed classifier is very robust to the weather conditions changes as well.

\begin{table*}[htb]
    \caption{Results of parking slot occupancy classification on CNRPark-EXT dataset - parking lot generalization.}
    \label{tab:classification_AUC_results_CNRPark-EXT}
    \footnotesize
    \begin{tabularx}{\textwidth}{LLLLL}
    \hline
    \rowcolor{LightCyan}
    \textbf{Method} 		& \textbf{Training subset} 		& \textbf{Testing subset} 		& \textbf{Testing accuracy} 	& \textbf{AUC} \\ 
	\hline
        
    Ours 						& \multirow{3}{2.5cm}{CNRPark} 					& \multirow{3}{2.5cm}{CNRPark-EXT TEST} 	& \textbf{97.66\%}	& \textbf{0.9969}	\\
    mAlexNet 				 	& 			 									& 											& 93.52\% 	& 0.9838	\\
    AlexNet 				 	& 			 									&  											& 93.63\% 	& 0.9877	\\
    
    \hline
    
    Ours 						& \multirow{3}{2.5cm}{CNRPark+EXT TRAIN C1-C8}	& \multirow{3}{2.5cm}{CNRPark-EXT TEST} 	& \textbf{99.34\%}	& \textbf{0.9994}			\\
    mAlexNet 				 	&  												& 											& 95.88\% 	& 0.9937	\\
    AlexNet 					& 												& 											& 96.85\% 	& 0.9957	\\
    
    \hline
	
	Ours 						& \multirow{4}{2.5cm}{CNRPark+EXT TRAIN}		& \multirow{4}{2.5cm}{CNRPark-EXT TEST} 	& \textbf{99.67\%} 	& \textbf{0.9981}			\\
    mAlexNet 					& 												& 											& 97.71\% 	& 0.9967	\\
    AlexNet 				 	& 		 										& 										 	& 98.00\% 	& 0.9974	\\    
	CarNet 						&												&											& 98.11\%	& N/A		\\
    \hline
    \end{tabularx}
\end{table*}

\begin{table*}[htb]
    \caption{Results of parking slot occupancy classification on CNRPark-EXT dataset - viewpoint generalization.}
    \label{tab:classification_AUC_results_CNRPark-EXT_viewpoint}
    \footnotesize
    \begin{tabularx}{\textwidth}{LLLL}
	\hline
    \rowcolor{LightCyan}
    & & \multicolumn{2}{c}{\textbf{Testing accuracy}} \\
	\rowcolor{LightCyan}
    \multirow{-2}{*}{\textbf{Training subset}} & \multirow{-2}{*}{\textbf{Testing subset}} & \mbox{\textbf{Ours}} & \mbox{\textbf{mAlexNet}} \\
    \hline
    \multirow{9}{2.5cm}{CNRPark-EXT C1}		& CNRPark-EXT C1	& -					& - 		\\
    										& CNRPark-EXT C2	& \textbf{99.19\%}	& 94.85\%	\\
    										& CNRPark-EXT C3	& \textbf{97.28\%}	& 93.11\%	\\
    										& CNRPark-EXT C4	& \textbf{97.89\%}	& 96.00\%	\\
    										& CNRPark-EXT C5	& \textbf{97.60\%}	& 95.91\%	\\
    										& CNRPark-EXT C6	& \textbf{97.25\%}	& 95.61\%	\\
    										& CNRPark-EXT C7	& \textbf{96.99\%}	& 91.43\%	\\
    										& CNRPark-EXT C8	& \textbf{97.43\%}	& 94.61\%	\\
    										& CNRPark-EXT C9	& \textbf{95.75\%}	& 90.96\%	\\
    \hline
    \multirow{9}{2.5cm}{CNRPark-EXT C8}		& CNRPark-EXT C1	& \textbf{92.79\%}		& 92.39\%	\\
    										& CNRPark-EXT C2	& \textbf{99.99\%}		& 94.51\% 	\\
    										& CNRPark-EXT C3	& \textbf{97.01\%}		& 93.66\% 	\\
    										& CNRPark-EXT C4	& \textbf{98.29\%}		& 97.53\% 	\\
    										& CNRPark-EXT C5	& 97.49\%				& \textbf{97.93\%} 	\\
    										& CNRPark-EXT C6	& 96.72\%				& \textbf{97.68\%} 	\\
    										& CNRPark-EXT C7	& \textbf{95.55\%}		& 93.53\% 	\\
    										& CNRPark-EXT C8	& -						& - 		\\
    										& CNRPark-EXT C9	& \textbf{96.47\%}		& 94.65\% 	\\    									
    \hline
    \end{tabularx}
\end{table*}

\begin{table*}[htb]
    \caption{Results of parking slot occupancy classification on CNRPark-EXT dataset - weather generalization.}
    \label{tab:classification_AUC_results_CNRPark-EXT_weather}
    \footnotesize
    \begin{tabularx}{\textwidth}{XXXX}
	\hline
    \rowcolor{LightCyan}
    & & \multicolumn{2}{c}{\textbf{Testing accuracy}} \\
	\rowcolor{LightCyan}
    \multirow{-2}{*}{\textbf{Training subset}} & \multirow{-2}{*}{\textbf{Testing subset}} & \mbox{\textbf{Ours}} & \mbox{\textbf{mAlexNet}} \\
    \hline
    \multirow{2}{2.5cm}{CNRPark-EXT SUNNY}			& OVERCAST		& \textbf{99.69\%}		& 97.80\% 		\\
    												& RAINY			& \textbf{99.02\%}		& 95.79\% 		\\
    									
    \hline
    \multirow{2}{2.5cm}{CNRPark-EXT OVERCAST}		& SUNNY			& \textbf{98.95\%}		& 91.63\% 		\\
    												& RAINY			& \textbf{98.11\%}		& 94.68\% 		\\
    											
    \hline
    \multirow{2}{2.5cm}{CNRPark-EXT RAINY}			& SUNNY			& \textbf{98.54\%}		& 93.53\% 		\\
    												& OVERCAST		& \textbf{99.39\%}		& 98.27\%		\\   														
    \hline
    \end{tabularx}
\end{table*}

\section{Conclusions}
\label{sec:Conclusions}

In this paper, vision-based algorithm called APSD-OC is proposed. APSD-OC automatically detects parking slots and classifies each parking slot as occupied or free. As such it contains two main parts. In the first part, the locations of parking slots in the input image are determined. Hereby, vehicles are detected in a series of input images using YOLOv5. After that, the detections centers are transformed to bird's eye view using a homography matrix which is obtained by a CNN. The transformed detections are clustered using DBSCAN algorithm. The resulting centers are filtered and are projected back to the original view where they correspond to the locations of parking slots. In the second part of the algorithm, the detected parking slots are classified as occupied or vacant using a ResNet34 based classifier.

The proposed APSD-OC algorithm is evaluated on two publicly available datasets: PKLot and CNRPark-EXT. The evaluation is carried on by analyzing parking slot detection and parking slot occupancy classification performance of the proposed APSD-OC algorithm. The obtained results show that detection precision and recall goes well beyond 90\% as more input images are used in the detection procedure on both datasets. The proposed algorithm is robust to the presence of parking violations and passing vehicles which are often appearing in images of the PKLot dataset. The trained deep classifier shows high accuracy, obtaining AUC over 0.99 for different combinations of training and testing subsets in the case of PKLot dataset thus significantly outperforming CarNet and mAlexNet. The same can be observed in the case of CNRPark-EXT dataset, proving high classifier robustness to viewpoint change and weather conditions.

Our future work will include estimation of the number of regular parking slots from the shape of the vehicle detections distribution. Apart from that, we will try to utilize the fact that parking slots have a certain spatial relationship to obtain even more accurate parking slot detection.

\bibliography{bibliography}

\end{document}